\pdfoutput=1
\documentclass[11pt]{article}

\usepackage[table,dvipsnames]{xcolor}
\usepackage{acl}

\usepackage{times}
\usepackage{latexsym}
\usepackage{inconsolata}
\usepackage{amsfonts}
\usepackage{soul}
\usepackage{colortbl}
\usepackage{multirow}
\usepackage{xspace}
\usepackage{amsmath}
\usepackage{graphicx}
\usepackage{booktabs}
\usepackage{tcolorbox}
\usepackage{caption}
\usepackage{subfigure}
\definecolor{lightred}{RGB}{255,179,179}

\newcommand{\gpt}{GPT-3.5\xspace}
\newcommand{\gem}{Gemini\xspace}
\newcommand{\llama}{LLaMA\xspace}

\newcommand{\unisample}{$\mathcal{U}$\xspace}
\newcommand{\ransample}{$\mathcal{R}$\xspace}

\usepackage[T1]{fontenc}
\usepackage[utf8]{inputenc}

\usepackage{microtype}

\usepackage{inconsolata}

\title{Linguistic Blind Spots of Large Language Models
}

\author{Jiali Cheng \quad Hadi Amiri\\
  University of Massachusetts Lowell \\
  \texttt{\{jiali\_cheng, hadi\_amiri\}@uml.edu} \\}

\begin{document}
\maketitle
\begin{abstract}
% the focus ins on identification of linguistic units and adherence to linguistic rules

Large language models (LLMs) 
% have made successful transformed Natural Language Processing (NLP) and 
are the foundation of many AI applications today. However, despite their remarkable proficiency in generating coherent text, questions linger regarding their 
% depth of linguistic understanding. 
ability to perform fine-grained linguistic annotation tasks, such as detecting nouns or verbs, or identifying more complex syntactic structures like clauses in input texts. These tasks require precise syntactic and semantic understanding of input text, and when LLMs underperform on specific linguistic structures, it raises concerns about their reliability for detailed linguistic analysis and whether their (even correct) outputs truly reflect an understanding of the inputs.
In this paper, we empirically study the performance of recent LLMs on fine-grained linguistic annotation tasks. 
% by asking linguistic-heavy question, such as recognizing T-units. 
Through a series of experiments, we find that recent LLMs show limited efficacy in addressing linguistic queries and often struggle with linguistically complex inputs. We show that the most capable LLM (Llama3-70b) makes notable errors in detecting linguistic structures, such as misidentifying embedded clauses, failing to recognize verb phrases, and confusing complex nominals with clauses. 
% hallucinating or omitting information, disregarding instructions, and misspelling common words. 
% In addition, on 33\% of test samples that contain repeating POS tags, the model failed to retrieves all POS tags of the same type in a sample. 33\%
Our results provide insights to inform future advancements in LLM design and development.\looseness-1 %alignment. 

\end{abstract}

\section{Introduction}
Large Language Models (LLMs) have revolutionized NLP by achieving remarkable performance on a wide range of tasks and applications, including 
zero-shot inference~\citep{weller-etal-2020-learning,NEURIPS2020_1457c0d6}; 
solving math problems~\citep{wei2022chain}; 
representing human emotions~\citep{li2024emotion}; and
serving as planners~\citep{pmlr-v162-huang22a},
conversational agents~\citep{ouyang2022training}, 
or text-to-code convertors~\citep{sun-etal-2023-battle}.
% or even learners of new tools~\citep{schick2023toolformer}. 
Nevertheless, despite recent studies~\citep{shen-etal-2021-whats,yu2023white,chen2024sudden} aiming to understand Transformers~\citep{NIPS2017_3f5ee243} as the building block of LLMs,  
there is a lack of systematic evaluation of their ability in performing fine-grained linguistic annotation tasks.

% LLMs are predominantly employed as blackbox models, particularly in case of extremely large models such as GPTs~\citep{ouyang2022training} with billions of parameters. Furthermore, from the linguistic perspective, while LLMs show remarkable proficiency in processing and producing textual data, questions remain about their depth of linguistic capability.

% Recent evaluation of LLMs focus on 
% math ability~\citep{}, 
% coding ability~\citep{}, 
% planning~\citep{},
% multilingual ability~\citep{ahuja-etal-2023-mega}.
% There lacks a comprehensive evaluation of LLM's knowledge on linguistic.

\begin{figure}[t]
    \centering
    \subfigure[Penn Treebank]{\includegraphics[width=0.48\linewidth]{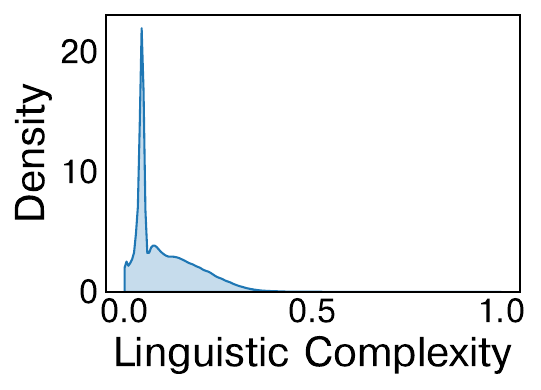} }
     \subfigure[CoNLL 2000]{\includegraphics[width=0.48\linewidth]{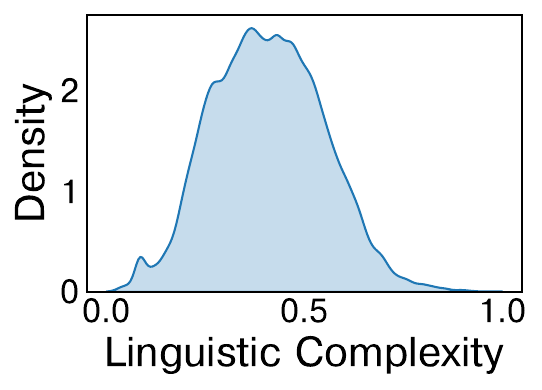} }
     \vspace{-10pt}
\caption{Distribution of linguistic complexity in two widely-used NLP datasets. The plots show \textbf{(a)}: a strong skew toward linguistically simple examples in the Penn Treebank and \textbf{(b)}: a concentration around moderate complexity in CoNLL 2000, which highlights an overrepresentation of easier or medium-difficulty samples in the datasets.}
\label{fig:dist_ling_ind}
\vspace{-5pt}
\end{figure}

Recent work studied LLMs from different linguistic perspectives, including grammar learning with small models~\citep{huebner-etal-2021-babyberta}, effect of pre-training on learning linguistic properties like the depth of parse tree or verb tense~\citep{alajrami-aletras-2022-pre}, the role of individual neurons in POS tagging and chunking tasks~\citep{durrani-etal-2020-analyzing}, and the effect of prompt design for detecting linguistic properties~\citep{blevins-etal-2023-prompting}. % A detailed description of related work is presented in Appendix~\ref{sec:related}.
However, existing evaluations are based on NLP datasets where linguistically ``easy'' examples (see Section~\ref{sec:ling_complexity}) are overrepresented. For instance, Figure~\ref{fig:dist_ling_ind} shows histograms of the linguistic complexity of samples in two widely-used NLP datasets: Penn Treebank~\citep{marcus-etal-1993-building} and CoNLL 2000~\citep{tjong-kim-sang-buchholz-2000-introduction}. The skewed distribution toward linguistically easy or medium examples can artificially inflate performance on NLP tasks\footnote{This phenomenon has challenged the NLP community across natural language inference (NLI), POS tagging, and parsing tasks, where models show human-level performance, while lacking cognitive ability to address these tasks. For example, recent work by~\citet{sinha-etal-2021-unnatural} shows that BERT is invariant to random word order permutation in case of NLI, which can be attributed to the high prevalence of linguistically easy samples in NLI datasets~\citep{elgaar-amiri-2023-ling}.} and prevent true evaluation of models in NLP.
We mitigate this bias by reducing the effect of overrepresented examples, i.e., categorizing samples based on their linguistic complexity and uniformly sampling data from distinct groups for a more reliable assessment.

We investigate the following research questions: 
(1): {\em how accurately can recent LLMs detect complex linguistic structures in input text?} 
(2): {\em which linguistic structures represent the blind spots of recent LLMs--meaning the most challenging for them?} 
(3): {\em how does the performance of LLMs vary across different levels of linguistic complexity of inputs?} 
We answer these questions by designing an empirical study
% and different prompting strategies
for LLMs. 
% Our prompting strategies are designed to assess the linguistic capabilities of recent LLMs. 
% Our prompting strategies include plain prompting, which queries LLMs to detect specific linguistic structures within text; chain-of-thought prompting, which employs instructions to query LLMs to complete a linguistic task; and structured-prompting~\citep{blevins-etal-2023-prompting}, which sequentially presents linguistic queries to LLMs following each prediction for partial inputs, see Section~\ref{sec:prompt}. 
The contributions of this paper are in examining recent LLMs's ability to detect specific linguistic structures across varying levels of linguistic complexity, providing meaningful insights into their limitations and biases, and highlighting potential avenues for future improvements.
% discussion on the future development of LLMs.
% \item The ground truth linguistic entity mentions and predicted outputs by \gpt can be used for other purposes, such as forming a recognition task

%Extensive 
Experimental results show that recent LLMs have limited efficacy in addressing linguistic queries, particularly struggling with complex linguistic structures such as complex nominals and T-units. In particular, Llama3-70b and GPT-3.5 are the most capable models among evaluated LLMs, while still making mistakes on simple linguistic queries. In addition, the performance of all evaluated LLMs often substantially fluctuates as sample complexity varies.
%
% In addition, we show that the most capable LLM in our experiment (\gpt) makes notable errors in linguistic annotations. 
% such as hallucinating or omitting information, disregarding instructions, and misspelling common words. 
% They struggle to understand some linguistic tasks and contexts, such as sentence chunking and CoNLL 2000 format. They are poorly aligned under many cases, failing to strictly follow human instructions. They can generate significant amount of misleading or inaccurate output. These results raise the question that if the current LLMs really understand human language, despite their great success on a wide range of tasks. Our analysis sheds light on how we design, train and align future language models. 

% We sample examples uniformly across the difficulty spectrum to get a fair evaluation.

\section{Background}
\label{sec:ling_complexity}

\paragraph{Linguistic Complexity:} quantifies the variability and sophistication in productive vocabulary, grammatical structures, and fluency in text data. It has been extensively investigated in psycholinguistics literature~\citep{wolfquintero1998,zareva2005relationship,Lu2010,housen2019multiple,biber2020investigating}; 
% bulte2012defining
and examined in quantifying language proficiency~\citep{yannakoudakis-etal-2011-new,Lu2012}, 
readability assessment and text simplification~\citep{feng-etal-2009-cognitively,xu-etal-2015-problems,xia-etal-2016-text,lee-etal-2021-pushing}, and improving NLP tasks~\citep{wei-etal-2021-linguistic}.
% \textbf{?}

% In particular, \citet{wolfquintero1998},~\citet{bulte2012defining}~and~\citet{housen2019multiple} developed comprehensive taxonomies of different complexity constructs in English language. 

\paragraph{Lexical Complexity:} is concerned with lexical \textit{density}, \textit{sophistication}, and \textit{variation}. 
Lexical density is often quantified by 
% the ratio of the number of \textit{open-class} words to the total number of words in a given text. Texts with higher lexical density are expected to be more complex as they contain larger 
the extent of information-carrying words in inputs.
Lexical sophistication measures the proportion of \textit{sophisticated} or infrequent words in texts. 
Lexical Variation refers to the diversity of vocabulary in text. Examples include type-token ratio~\citep{templin1957certain} and its variations including {\em D-measure}~\citep{malvern2004lexical}, which 
determines lexical variation of text by finding 
the curve that best fits the actual relationship between types and tokens in input text.

\paragraph{Syntactic Complexity:} determines variability and sophistication in grammatical structures. A sentence like ``\textit{the mouse ate the cheese}'' can be converted to its well-formed yet complex counterpart ``\textit{the mouse the cat the dog bit chased ate the cheese},'' which forces readers to suspend their partial understanding of the sentence by encountering subordinate clauses that substantially increase the cognitive load of the sentence. Syntactic complexity measures the length of production units at the clausal, sentential, or T-unit levels; the amount of subordination, e.g. number of clauses per T-unit; % or dependent clauses per clause
the amount of coordination, e.g. number of coordinate phrases per clause or T-unit; and the range of surface and particular syntactic and morphological structures, e.g. frequency and variety of tensed forms~\citep{wolfquintero1998,ortega2003syntactic}. 

\begin{figure*}[t]
    \centering
    \includegraphics[width=.8\textwidth]{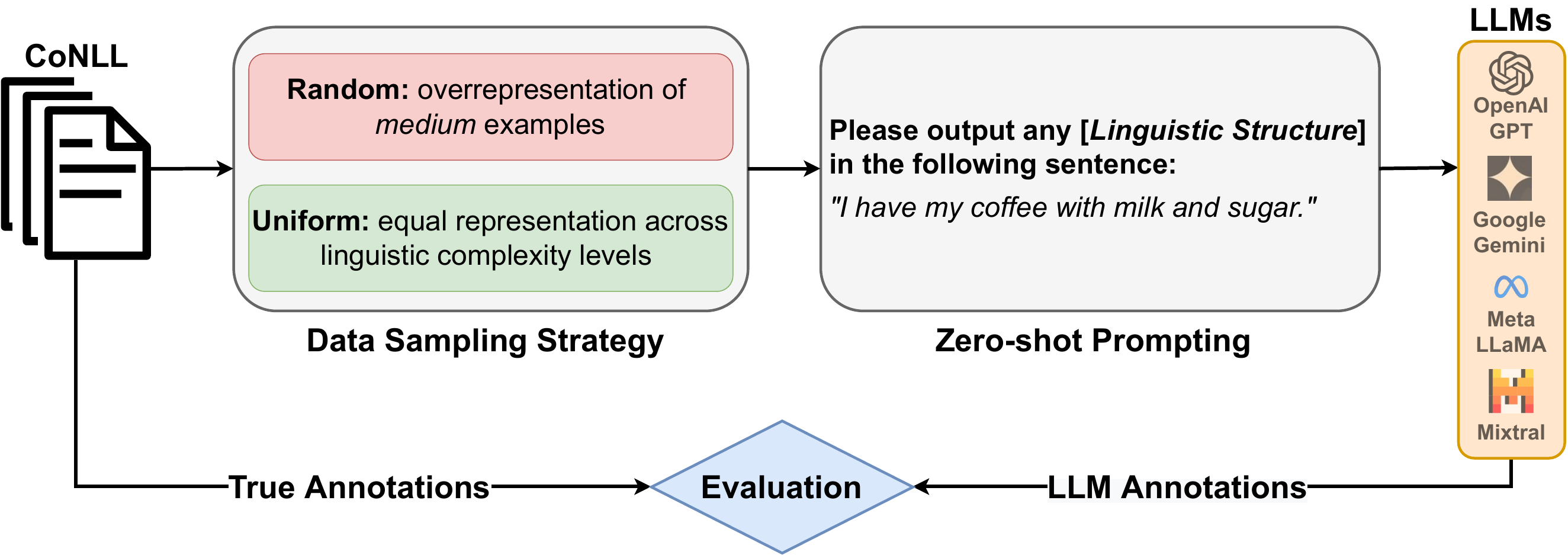}
    \caption{Workflow for finding linguistic blind spots of LLMs. As illustrated in Appendix~\ref{sec:gpt_knowledge}, GPT and other LLMs have good knowledge of our target tasks and the relevant terminology used in the prompts. [\textit{Linguistic Structure}] in the prompts indicate any of the lexical or syntactic structures listed in Appendix~\ref{sec:app}.}
\label{fig:flow}
\end{figure*}

\paragraph{Linguistic Knowledge of LLMs}
\citet{blevins-etal-2023-prompting} designed {\em structured} prompting to assess the linguistic capabilities of LLMs. 
They provided each LLM with fully labeled demonstrations, and a query sentence and its partially tagged version. Each predicted label was appended to the partially tagged query along with the next word to iteratively tag the full query. They found that GPT-3.5 is robust to arbitrary label selections and ignores labels conflicting with its prior knowledge, indicating that the models can learn general linguistic knowledge during pre-training, rather than simply memorizing the data.
\citet{alajrami-aletras-2022-pre} empirically compared linguistically-motivated (e.g. masked language modeling~\citep{devlin-etal-2019-bert}) and non-linguistically motivated (e.g. masked first character prediction~\citep{yamaguchi-etal-2021-frustratingly}) pre-training objectives for BERT on linguistic probing tasks~\citep{linzen-etal-2016-assessing,warstadt-etal-2020-blimp-benchmark}. They found the two objectives achieve similar performance.
% , demanding deeper analysis of LLMs and pre-training objectives.
%
\citet{clark-etal-2019-bert} showed that attention heads in transformers attend to boundary tokens, positional offsets, and whole sentence; while \citet{voita-etal-2019-analyzing} showed that attention heads mainly handle positions, syntax, and rare words. 
% Pruning heads that do not play the above roles barely hurts the performance on several tasks. 
%
\citet{durrani-etal-2020-analyzing} compared linguistic knowledge learned by LMs at neuron level. They narrowed down 
% linguistic-related 
neurons to a specific subset, located in lower hidden layers for lexical knowledge and in higher layers for semantic knowledge. 
%
%for ELMo~\citep{peters-etal-2018-deep} and XLNet~\citep{NEURIPS2019_dc6a7e65} are clustered together. While BERT~\citep{devlin-etal-2019-bert} have more scattered task-specific neurons. 
Finally, \citet{sharma-etal-2023-learning} found that learning non-linguistic knowledge (e.g. numerical skills) sacrifices the linguistic knowledge of LLMs, and \citet{ettinger-2020-bert} found that BERT underperforms on commonsense, pragmatic inference, and negation tasks.

\section{Finding Linguistic Blind Spots}\label{sec:method}

% \paragraph{Linguistic Probing Tasks and Complexity}
We evaluate LLMs on recognizing specific linguistic structures (see below).
% \subsection{Linguistic structures}
% We introduce a list of common linguistic structures (structures), and prompt \gpt to recognize them in the input sentences in a zero-shot fashion.
% \paragraph{Word-level structures} We test if LLMs can correctly identify different types of word-level structures $\mathcal{C}_W$ in text. The structures that we consider are: nouns (NN), verbs (VB), adjectives (JJ), adverbs (RB), preposition/subordinate (IN), coordinating conjunctions (CC), determinants (DT).
% \paragraph{Phrase-level structures} We test if LLMs can correctly identify different types of phrase-level structures $\mathcal{C}_P$, including noun phrases (NP), verb phrases (VP), adjective phrase (ADJP), adverb phrase (ADVP), preposition phrase (PP), conjunction phrases (CONJP), coordination phrases (CP), quantitative phrase (QP), and complex nominals (CN).
% \paragraph{Sentence-level structures} We test if LLMs can correctly recognize sentence-level structures $\mathcal{C}_S$, including clauses (C), dependent clauses (DC), fragment clause (FC), t-units (T), and complex t-unit (CT) in input text.
% \paragraph{Linguistic Complexity}
% \label{sec:ling_ind}
% We collect a diverse collection of linguistic indices from related works~\citep{Lu2012,Lu2010,lee-etal-2021-pushing}, shown in Appendix Table~\ref{tab:ling_ind}. We compare \gpt's performance as the linguistic indices vary.
% Linguistic index is a measurement of the difficulty and diversity of a piece of text. In this work, 
For this purpose, we use gold linguistic annotations, lexical complexity analyzer from~\citep{Lu2012}, and syntactic complexity analyzer from~\citep{Lu2012} 
% number of different POS and syntactic tags provided in the ground truth 
to quantify linguistic complexity of samples. We note that the estimations provided by these tools have perfect agreement (based on Cohen's Kappa) with estimations provided by more recent linguistic complexity analysis tools~\citep{lee-etal-2021-pushing,lee-lee-2023-lftk}.
% We normalize the tag counts for each tag type, denoted as $\mathbb{LI}_t$. The overall linguistic complexity of each sample ($\Bar{\mathbb{LI}}$) is then determined by averaging its normalized tag counts. 
% We can also average the linguistic indices by the level of the tags, where $\mathbb{LI}_P$ and $\mathbb{LI}_S$ denote average POS-level and syntax-level indices, respectively.
% We note that existing linguistic complexity measurements are mostly derived from the counts, and have their own biases towards specific domain. For example, \citet{Lu2010} has less coverage on adjectives and adverbs and \citet{lee-lee-2023-lftk} relied on trained pipelines which may be misleading. Moreover, the above estimations have substantial agreement (based on Cohen's Kapps) with estimations provided by dedicated linguistic complexity analysis tools~\citep{lee-etal-2021-pushing,lee-lee-2023-lftk,Lu2010}. 
% We use the counting approach for comprehensive coverage over structures to ensure a thorough evaluation. We also show in experiments that using counting approach leads to a better separation of token-, phrase- and sentence-level complexity measures for a fairer evaluation against texts of various levels of linguistic complexity

\paragraph{Linguistic Structures:}
we consider different levels of granularity:
\textbf{word-level structures} like nouns, verbs, adjectives, adverbs, prepositions, conjunctions, numerals, determiners, punctuation, particles, and words that cannot be assigned a part-of-speech (POS) tag; 
\textbf{phrase-level structures} like noun phrases (NP), verb phrases (VP), adjective phrases (ADJP), adverb phrases (ADVP), conjunction phrases (CONJP), complex nominals (CN); and 
\textbf{sentence-level structures} like clauses (C), dependent clauses (DT), T-units (T), and complex T-units (CT).
Appendix~\ref{sec:app} lists these structures.

\paragraph{Data Sampling Strategy}
The overrepresentation of easy and medium examples shown in Figure~\ref{fig:dist_ling_ind} suggests that the linguistic capability of LLMs may have been overestimated in existing literature~\citep{blevins-etal-2023-prompting,yang-tu-2022-bottom,shen-etal-2018-straight}. For fair evaluation across the linguistic complexity spectrum, we divide samples into eight groups of increasing linguistic complexity, determined using~\citep{Lu2010,Lu2012}, and uniformly at random sample from each group, leading to a total number of 8$\times$125 = 1k samples, denoted as \unisample. For comparison, we also randomly select 1k samples from the dataset, which shows similar distribution to the original distribution, denoted as \ransample.
% with overrepresentations of medium examples, denoted as \ransample.

\paragraph{Prompting Strategies:} %\label{sec:prompt}
% we employ different prompting strategies to evaluate LLMs in identifying linguistic structures in input text:
% % to probe LLMs linguistic %knowledge
% % capabilities, illustrated in Figure~\ref{fig:flow}: 
% \textbf{zero-shot prompting} tasks the LLM with directly identifying individual linguistic structures in input text in a question-answering format, see zero-shot prompting in Figure~\ref{fig:flow}; and  %
% \textbf{chain-of-thought (CoT) prompting}, which extends zero-shot prompting by prompting LLMs to generate intermediate reasoning steps with the phrase\emph{let's think step-by-step}, Figure~\ref{fig:flow}. 
% Given the prevalent use of current LLMs as conversational agents, these prompting strategies align naturally with their operational context.
we use zero-shot prompting to assess LLMs' ability to identify individual linguistic structures in input text in a question-answering format, see Figure~\ref{fig:flow}. We also investigate other prompting techniques, such as manually optimizing instructions, chain-of-thought (CoT) prompting~\citep{wei2022chain} and structured prompting~\citep{blevins-etal-2023-prompting}. However, in a small scale experiment, the alternative approaches did not result in consistent performance improvement over the zero-shot approach. This could be because LLM's current pretraining does not fully capture the complex syntactic and semantic information of inputs required for fine-grained linguistic annotation. Instead, they might rely heavily on surface-level patterns, which limits the impact of more advanced prompting strategies.

\section{Experimental Setup}

% \paragraph{Dataset}
% We conduct the experiments on the PTB Release 3~\citep{10.5555/972470.972475}, since it is labeled by human experts with ground truth Part-of-Speech (POS) tags. The PTB corpus consists of documents from the following three sources: 
% \begin{itemize}
%     \itemsep-1pt
%     \item Wall Street Journal articles (WSJ), which contains news articles;
%     \item Brown Corpus dataset (Brown), which consists of multi-genera multi-source American English text. PTB includes a subset of the Brown corpus, such as biography and fictions;
%     \item Switchboard (SWBD), which consists of two-sided telephone conversations from multiple speakers~\citep{switchboard}. 
% \end{itemize}

\paragraph{Dataset \& Evaluation:}
We use the CoNLL 2000~\citep{tjong-kim-sang-buchholz-2000-introduction} subset of the Penn Treebank corpus~\citep{marcus-etal-1993-building} (Wall Street Journal (WSJ) sections 15, 16, 17, 18, 20), which provides ground truth POS tags and syntactic annotations. We use standard pre-processing to convert POS tags to Universal POS tags~\citep{blevins-etal-2023-prompting}. Following previous work~\citep{blevins-etal-2023-prompting}, we compute precision, recall, and F1 score for each sample, and average them across all samples to evaluate LLM performance in recognizing linguistic structures. 
% and clean the parsing trees~\citep{yang-tu-2022-bottom}. 
% We directly use the chunking labels provided by the CoNLL 2000 sentence chunking task.

% Figure~\ref{fig:dist_li} presents the difficulty distribution of the 1K samples, showing that our selection criterion results in a wide and uniform coverage on the difficulty spectrum. 
% The details of the 1K samples are shown Table~\ref{tab:corpus}. 
% For the mean linguistic index value less than 4, we divide it into bin of . Samples whose linguistic indices are greater than 0.4 are considered as hard examples. 
% We sample 1K sentences from the three domains with stratification based on their proportions, with 

% \begin{table}[h]
% \small
% \centering
% \begin{tabular}{l|ccc}
% \toprule
%     Entity       & WSJ & Brown & SWBD \\
%     \midrule
%     Domain       & News & Multi-genera text & Conversation \\
%                  % &      & fiction, etc & \\
%     Proportion   & 0.32 & 0.31 & 0.37 \\
%     \# sentences & 320 & 310 & 370 \\
% \bottomrule
%   \end{tabular}
%   \caption{Corpus statistics.}
%   \label{tab:corpus}
% \end{table}

\paragraph{Large Language Models:}
We use several robust LLMs including 
\gpt~\citep{ouyang2022training} (\texttt{gpt-3.5-0613}), 
Gemini-Pro 1.0~\citep{team2023gemini},
Llama3 (7B, 13B, 70B)~\citep{touvron2023llama},
Llama2 (7B, 13B, 70B), and 
Mistral (7B, 8x7B)~\citep{jiang2023mistral,jiang2024mixtral}.\looseness-1

\begin{table}[t]
\small
\centering
\begin{tabular}{l|c|ccc}
\toprule
LLM & Sampling & P & R & F1 \\
\midrule
Llama3-70b   & \ransample & \textbf{31.3}    &    \textbf{30.8} &    \textbf{29.2} \\
Llama3-70b   & \unisample & \textbf{28.2}    &    \textbf{27.5} &    \textbf{26.1} \\
\midrule
Llama3-8b    & \ransample & 24.0    &    26.8 &    23.2 \\
Llama3-8b    & \unisample & 21.8    &    24.0 &    20.8 \\
\midrule
GPT-3.5      & \ransample & 21.6    &    26.1 &    21.2 \\
GPT-3.5      & \unisample & 20.4    &    23.5 &    19.5 \\
\midrule
Llama2-70b   & \ransample & 13.4    &    21.4 &    14.7 \\
Llama2-70b   & \unisample & 11.8    &    18.4 &    12.8 \\
\midrule
Mixtral-8x7b & \ransample & 11.0    &    25.4 &    13.0 \\
Mixtral-8x7b & \unisample & 10.2    &    22.5 &    11.8 \\
\midrule
Mistral-7b   & \ransample & 7.4     &    15.2 &     8.0 \\
Mistral-7b   & \unisample & 6.9     &    13.5 &     7.5 \\
\midrule
Llama2-7b    & \ransample & 7.0     &     9.8 &     7.4 \\
Llama2-7b    & \unisample & 7.0     &     9.7 &     7.3 \\
\midrule
Gemini       & \ransample & 1.2     &     1.2 &     1.0 \\
Gemini       & \unisample & 1.2     &     1.2 &     1.1 \\

% \midrule
%   LLaMA3-13B &      \ransample &       14.4 &    18.8 &    14.3 \\
%   LLaMA3-13B &      \unisample &       13.2 &    16.6 &    12.8 \\
% \midrule
%   LLaMA2-13B &      \ransample &       14.4 &    18.8 &    14.3 \\
%   LLaMA2-13B &      \unisample &       13.2 &    16.6 &    12.8 \\
% \midrule
%  Vicuna2-13B &      \ransample &       13.1 &    21.6 &    12.6 \\
%  Vicuna2-13B &      \unisample &       12.6 &    19.7 &    11.9 \\
% \midrule
%    LLaMA-30B &      \ransample &        5.2 &    30.3 &     8.0 \\
%    LLaMA-30B &      \unisample &        5.0 &    28.2 &     7.6 \\
% \midrule
%   Vicuna-33B &      \ransample &        7.2 &    28.7 &     9.1 \\
%   Vicuna-33B &      \unisample &        6.9 &    26.4 &     8.7 \\
% \midrule
% Average & \ransample & 12.0 & 22.3 & 12.6 \\
% Average & \unisample & 11.2 & 20.1 & 11.7 \\
\bottomrule
\end{tabular}
    % \cmidrule{2-5}
    % GPT-3.5  & \ransample &  21.6 &    26.1 &    21.2 \\
    %          & \unisample &  20.4 &    23.5 &    19.5 \\
    %          \cmidrule{2-5}
    % LLaMA-2  & \ransample &  14.4 &    18.8 &    14.3 \\
    %          & \unisample &  13.2 &    16.6 &    12.8 \\
    %          \cmidrule{2-5}
    % Vicuna-2 & \ransample &  13.1 &    21.6 &    12.6 \\
    %          & \unisample &  12.6 &    19.7 &    11.9 \\
    % Chunking & 0.36 & 0.34 & 0.32 & 0.06 & 0.12 & 0.06 \\ 
    % Parsing  & 0.20 & 0.21 & 0.19 & 0.03 & 0.04 & 0.03 \\
  \caption{Average performance in identifying linguistic structures. We compute precision, recall, and F1 for each sample, and average them across all samples to assess LLM performance in detecting linguistic structures.\looseness-1}
  \label{tab:main_qp}
\end{table}

\section{Main Results}
% \subsection{Linguistic characteristics of PTB corpus}
% With the linguistic indices $\mathbb{LI}_c$, we compute their pairwise Pearson correlation as shown in Figure~\ref{fig:corr}.

% \begin{figure*}[h]
%     \centering
%     \includegraphics[width=0.8\textwidth]{figure/corr_indices.pdf}
%     \caption{Pearson correlation between linguistic indices.}
% \label{fig:corr}
% \end{figure*}

% For each sentence, we compute 42 linguistic indices of different granularity, including lexical, sytactic, and semantic, shown in Table~\ref{tab:ling_ind}. Fig.~\ref{fig:corr} in Appendix~\ref{sec:app} illustrates the correlation between these indices.

% With QP, \gpt achieves comparable precision recall on syntactic chunking, reflecting that it can retrieve some correct structures, but often can't correctly detect all structures. see Table~\ref{tab:main_structure} in Appendix~\ref{sec:more_result} for fine-grained performance of these LLMs across different POS tag categories. 

% Among the three flavors of LLMs, \gpt is the best one to be aware of the difference between word-level, phrase-level and sentence-level linguistic structures. While \vic and \flan both perform much worse than \gpt with close to zero performance in general.

\subsection{Deficient Linguistic Performance of LLMs}
Tables~\ref{tab:main_qp} 
% and Tables~\ref{tab:app_qp}--\ref{tab:app_ipsp} in Appendix~\ref{sec:more_result} 
show significant performance differences between LLMs when tasked with identifying linguistic structures across different sampling strategies.
Despite outperforming other LLMs by a large margin, Llama3-70b, Llama3-8b, and \gpt have considerably low performance in identifying linguistic structures. Among the evaluated LLMs, Llama3-70b performs the best, with average precision, recall, and F1 score of 31.3, 30.8, and 29.2 on randomly selected samples (\ransample), and 28.2, 27.5, and 26.1 on uniformly selected samples (\unisample). However, these results are substantially lower than that of traditional models with significantly smaller sizes~\citep{manning-etal-2014-stanford}. 

In addition, \gem, Llama-2 and Mistral show poor performance across all settings, indicating that many linguistic structures are indeed a blind spot for these LLMs. Larger scales of Llama2 and Mistral show slightly better performance, but still limited compared to \gpt and Llama3. These models often recognize the entire sentence as a phrase, can't distinguish between noun phrases (NPs) and verb phrases (VPs), and show poor performance in detecting clauses. Surprisingly, \gem lacks the ability to identify linguistic structures, with an average F1 score close to 0. Through manual analysis, we find that \gem often misinterprets linguistic queries with harmful content, see Section~\ref{sec:align}.

\subsection{Task Complexity}
We find all evaluated LLMs show stronger capability in detecting simpler linguistic structures (e.g. word-level) than more complex structures (e.g. sentence-level). Specifically, \gpt achieves an average F1 scores of 37.5 (\unisample) and 34.4 (\ransample) on word-level structures, but close to zero F1 on phrase-level and sentence-level structures, see Table~\ref{tab:per_ent}. For some complex structures including verb phrase (VP), complex nominal (CN), dependent clause (DC), T-unit (T), and complex T-unit (CT), all LLMs have close to zero F1 score. This might be because these complex structures require a model to detect simpler structures (e.g. POS tags) and build on them in a compositional manner to correctly identify the more complex ones. Our results show that LLMs can accomplish simpler linguistic tasks but fail to perform complex ones, which mainly require knowledge about compositionality.

\begin{table}[b!]
\small
\centering
\begin{tabular}{l|c|ccc}
\toprule
Structure & Sampling &  P & R & F1 \\
\midrule
\multicolumn{5}{l}{\emph{Word-level Structure}}\\
\midrule
  PUNC &   \ransample &      82.5 &    77.4 &    77.4 \\
  PUNC &  \unisample &       86.1 &    79.5 &    80.9 \\
  NOUN &   \ransample &      71.6 &    65.6 &    66.1 \\
  NOUN &  \unisample &       67.6 &    64.3 &    62.9 \\
  VERB &   \ransample &      61.4 &    61.4 &    55.9 \\
  VERB &  \unisample &       53.9 &    51.0 &    47.7 \\
   DET &   \ransample &      56.4 &    56.2 &    50.7 \\
   DET &  \unisample &       50.3 &    47.9 &    43.4 \\
   ADP &   \ransample &      48.7 &    60.1 &    50.2 \\
   ADP &  \unisample &       42.0 &    47.7 &    41.6 \\
   ADJ &   \ransample &      26.5 &    43.7 &    29.1 \\
   ADJ &  \unisample &       22.7 &    32.9 &    23.1 \\
   ADV &   \ransample &      25.1 &    37.0 &    26.6 \\
   ADV &  \unisample &       25.8 &    33.4 &    25.8 \\
  PRON &   \ransample &      18.0 &    35.1 &    20.1 \\
  PRON &  \unisample &       17.0 &    32.7 &    18.8 \\
   PRT &   \ransample &       8.5 &    34.7 &    12.7 \\
   PRT &  \unisample &        8.1 &    30.8 &    11.6 \\
  CONJ &   \ransample &      30.3 &    30.9 &    29.1 \\
  CONJ &  \unisample &       28.8 &    28.8 &    27.0 \\
   NUM &   \ransample &      31.5 &    29.8 &    29.7 \\
   NUM &  \unisample &       30.3 &    28.6 &    28.6 \\
     % X &   \ransample &       0.1 &     0.4 &     0.1 \\
     % X &  \unisample &        0.4 &     1.2 &     0.5 \\
Average & \ransample &       38.6 &    44.8 &    37.5 \\
Average & \unisample &	     36.2 &    40.2 &    34.4 \\
\midrule
\multicolumn{5}{l}{\emph{Phrase-level Structure}}\\
\midrule
  ADVP &   \ransample &       6.0 &    22.8 &     8.1 \\
  ADVP &  \unisample &        6.5 &    20.4 &     7.9 \\
    NP &   \ransample &      11.5 &    14.0 &    11.8 \\
    NP &  \unisample &       12.3 &    14.1 &    12.2 \\
  ADJP &   \ransample &       1.2 &     5.9 &     1.8 \\
  ADJP &  \unisample &        1.7 &     5.8 &     2.1 \\
    VP &   \ransample &       2.2 &     3.3 &     2.3 \\
    VP &  \unisample &        2.7 &     3.6 &     2.7 \\
 CONJP &   \ransample &       0.0 &     0.0 &     0.0 \\
 CONJP &  \unisample &        0.0 &     0.0 &     0.0 \\
    CN &   \ransample &       0.0 &     0.1 &     0.0 \\
    CN &  \unisample &        0.0 &     0.0 &     0.0 \\
Average & \ransample &        3.5 &     7.7 &     4.0 \\
Average & \unisample &	      3.9 &     7.4 &     4.2 \\
\midrule
\multicolumn{5}{l}{\emph{Sentence-level Structure}}\\
\midrule
     C &   \ransample &       0.1 &     0.3 &     0.1 \\
     C &  \unisample &        0.0 &     0.1 &     0.1 \\
    DC &   \ransample &       0.0 &     0.0 &     0.0 \\
    DC &  \unisample &        0.0 &     0.1 &     0.0 \\
     T &   \ransample &       0.0 &     0.0 &     0.0 \\
     T &  \unisample &        0.0 &     0.0 &     0.0 \\
    CT &   \ransample &       0.0 &     0.0 &     0.0 \\
    CT &  \unisample &        0.0 &     0.0 &     0.0 \\
Average & \ransample &        0.0 &     0.0 &     0.0 \\
Average & \unisample &	      0.0 &     0.0 &     0.0 \\
\bottomrule
\end{tabular}
\caption{Linguistic annotation performance of \gpt across different linguistic structure groups. We compute precision, recall, and F1 for each sample, and average them across all samples to assess LLM performance in detecting linguistic structures.}
\label{tab:per_ent}
\end{table}

\paragraph{\gpt Performance:} As shown in Table~\ref{tab:per_ent}, word-level structures such as nouns, verbs, and punctuation are generally better annotated by \gpt, while phrase-level and sentence-level structures, particularly verb phrases (VP), clauses (C), and complex T-units (CT), have significantly lower performance. These high-level structures are indeed blind spots for existing LLMs, due to their complexity and linguistic understanding required to accurately identify them.
Overall, \gpt tends to perform better on \ransample than on \unisample across most word-level and phrase-level structures. Specifically, on randomly selected samples, \gpt achieves average F1 scores of 37.5, 4.0, and 0.0 on word-level, phrase-level, and sentence-level structures respectively. On \unisample, \gpt achieves lower average F1 scores of 34.4 for word-level structures, 4.2 for phrase-level structures, and remains at 0.0 for sentence-level structures. These results indicates the model's relative strength in handling word-level structures but its significant limitation on more complex structures.

\subsection{Linguistic Complexity }
\paragraph{Performance Drop on Complexity-Balanced Samples:} We observe significant differences in LLMs' performances on \ransample and \unisample, as determined by a t-test at 95\% confidence interval. All evaluated LLMs (\gpt, \gem, Llama3, Llama2, Mistral) show significant decrease in performance on uniformly selected samples (\unisample) compared to randomly selected ones (\ransample). The only exceptions are \gem and Llama2-7B, which is likely due to their already low performance on both \ransample and \unisample. For \gpt, the performance drops from an F1 score of 21.2 to 19.5, with significant $p$-value of $1e$-$7$.
% from 57.5 to 54.5 with IP on POS tagging ($p$-value $= 7e-4$), 
% from 15.3 to 13.6 with IP on chunking ($p$-value $= 8e-3$), 
% from 75.9 to 73.0 with SP on POS tagging ($p$-value $= 4e-3$) and from 39.0 to 36.9 with SP on chunking ($p$-value $= 7e-3$).
We note that although the performance consistently and significantly decreases across models from R to U, the absolute drop is small to modest. This may be due to the already low overall performance ceiling on these tasks, where even small differences are meaningful; the models’ relative robustness to certain types of linguistic complexity, despite persistent weaknesses on edge cases and harder structures; or the prevalence of easier (word-level) structures compared to more complex (phrase- or sentence-level) ones in the set of linguistic structures we investigate.

\begin{figure*}[ht]
    \centering
    \includegraphics[width=0.98\textwidth]{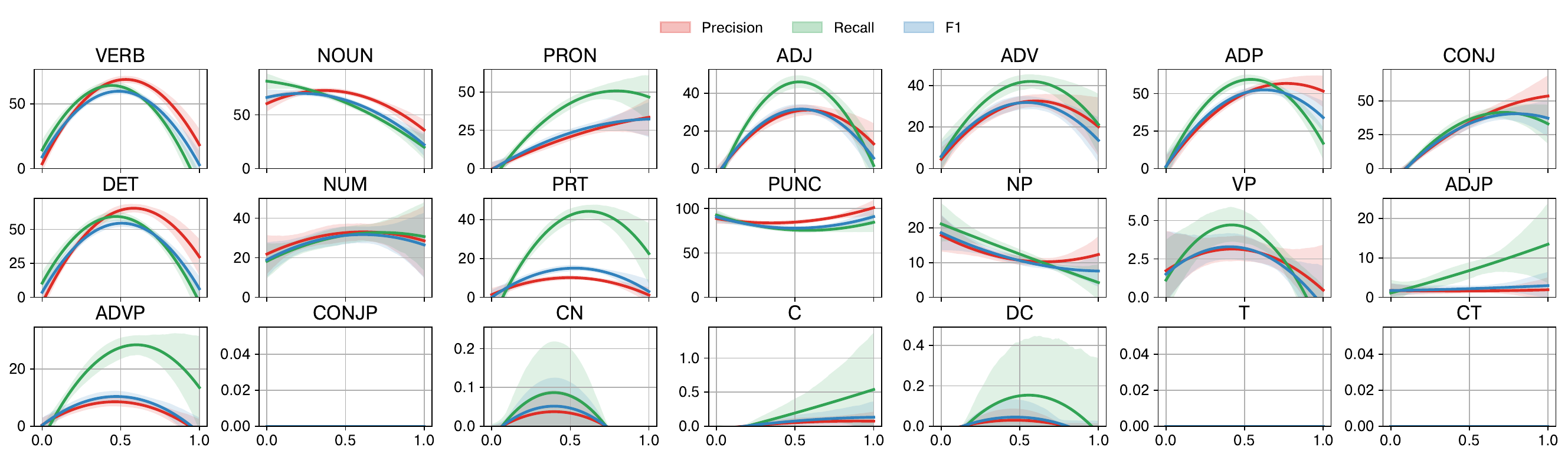}
    \caption{Performance of \gpt on texts of increasing linguistic complexity. \gpt achieves close to zero performance on CONJP, T, and CT. Figures~\ref{fig:qp_gemini}-\ref{fig:qp_mistral-7b} in Appendix~\ref{sec:more_result} show results of other LLMs.}
\label{fig:qp_gpt3.5}
\end{figure*}

\begin{figure}[t]
    \centering
    \includegraphics[width=0.8\linewidth]{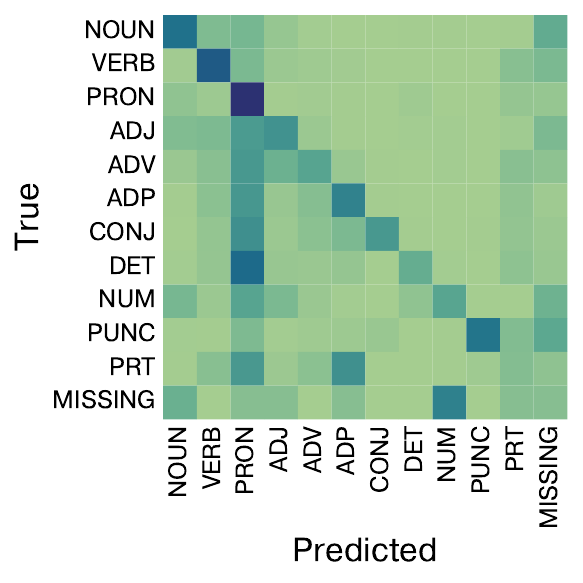}
    \vspace{-10pt}
    \caption{Confusion matrix of POS tagging on \gpt. Darker indicates larger value. Diagonal/off-diagonal elements represent correct/wrong predictions respectively.} %Results of other LLMs are shown in Figure~\ref{fig:conf_matrix_gemini}-\ref{fig:conf_matrix_llama2-7b}.}
\label{fig:conf_matrix_gpt3.5}
\vspace{-5pt}
\end{figure}

\paragraph{Linguistic Complexity Fluctuation:} We find that LLMs' performance fluctuate with increasing linguistic complexity of inputs, as shown in Figure~\ref{fig:qp_gpt3.5} for \gpt; see performance of other LLMs in Appendix~\ref{sec:more_result} Figures~\ref{fig:qp_gemini}--\ref{fig:qp_mistral-7b}. Specifically, the performance of \gpt improves initially but then declines on structures like verbs, nouns, pronouns, adjectives, and adverbs as linguistic complexity increases, with F1 scores ranging from 0 to 50. This suggest that expert-defined linguistic complexity~\citep{Lu2010,Lu2012} may not align with how LLMs view complexity, which is an underexplored topic. Interestingly, for other structures like punctuation (PUNC), we observe the opposite performance trend. This is likely due to the unique nature of these linguistic structures as punctuation marks typically follow more predictable and less complex rules compared to other linguistic structures like verbs or nouns.
In addition, performance trend vary substantially across different LLMs and scales. For instance, Llama3-70b consistently shows an inverted U-shaped ($\bigcap$) performance pattern, while Llama2-70b have unique trends on noun and punctuation, which indicate model-specific challenges with different linguistic structures.

\paragraph{POS Tag Errors in \gpt:}
Figure~\ref{fig:conf_matrix_gpt3.5} shows a confusion matrix that assess the POS tags generated by \gpt
% see performance of other LLMs in Appendix~\ref{sec:more_result}, Figures~\ref{fig:conf_matrix_gemini}--\ref{fig:conf_matrix_mixtral-7b}. 
Most of the errors stem from the model's failure to detect specific tags, denoted as ``MISSING.'' The higher occurrence of MISSING cases is likely due to the increased complexity and linguistic knowledge required for these tasks--the need to identify and label all instances of linguistic structures in inputs. In addition, \gpt often confuses different POS tags with pronouns. This could be because pronouns often appear in diverse contexts where their function can be easily confused with other POS tags, such as determiners or nouns. In addition, \gpt (and other LLMs) tend to rely on surface-level patterns rather than deep linguistic understanding. Pronouns frequently co-occur within sentences, and the model may overgeneralize their patterns to other words.

% In contrast, errors incurred for the QP prompting strategy are more varied and distributed across pronouns (PRON), prepositions (ADP), numerals (NUM), particles (PRT) and others such as foreign words, typos, abbreviations, or any other word that cannot be clearly assigned a universal POS category). We conjecture that the accurate detection of these tags often relies heavily on the surrounding context, as they can have multiple meanings or functions depending on the context in which they appear. This contextual variability makes their individual, almost contextless, detection a challenging task. 

% In addition, \gpt with IP strategy tends to miss particles (PRT), conjunctions (CONJ), and others (mentioned above), while often assigns prevalent tags such as nouns (NOUN), prepositions (ADP) or determiners (DET) when generating wrong tags. With SP, however, \gpt makes fewer mistakes. 
%
% Manual analysis of the data shows that many errors appear in the form of missing tags, particularly under the %QP and 
% the IP strategy. 
% , which usually happen to punctuation.

\begin{table}[t]
\setlength{\tabcolsep}{3pt}
\small
\centering
\begin{tabular}{l|ccccccc}
\toprule
    Entity & NOUN & VERB & ADJ & ADV & ADP & CONJ & DET \\
    \midrule
    \# Dup.  & 334 & 370 & 93 & 156 & 526 & 400 & 635 \\
    \# Succ. & 0 & 1 & 0 & 0 & 5 & 1 & 2 \\
\bottomrule
  \end{tabular}
  \caption{\gpt performance on samples that contain multiple instances of the same linguistic structure. Dup. indicates number of such texts (out of 1K) for each structure and Succ. indicates cases where {\em all} instances of the same POS tag are retrieved.}
  \label{tab:incomp}
\end{table}

\subsection{Multiple Structures and False Positives}
When a samples contains multiple occurrences of the same linguistic structure, such as nouns, LLMs often struggle to retrieve all instances of of those structures. 
% For example, if a sentence contains 4 nouns, detecting 3 of them is an incomplete retrieval. 
Table~\ref{tab:incomp} shows that \gpt consistently fails to identify {\em all} nouns in any of the 334 samples containing more than one noun. This limitations extends beyond open-class words to closed-class tags such as prepositions (ADP), conjunctions (CONJ) and even determiners (DET). 
% We notice that although determiners are a fixed set of simple words, \gpt only successfully retrieves all of the determiners two times out of 635 cases.

\begin{figure}[t]
    \centering
    \includegraphics[width=.98\linewidth]{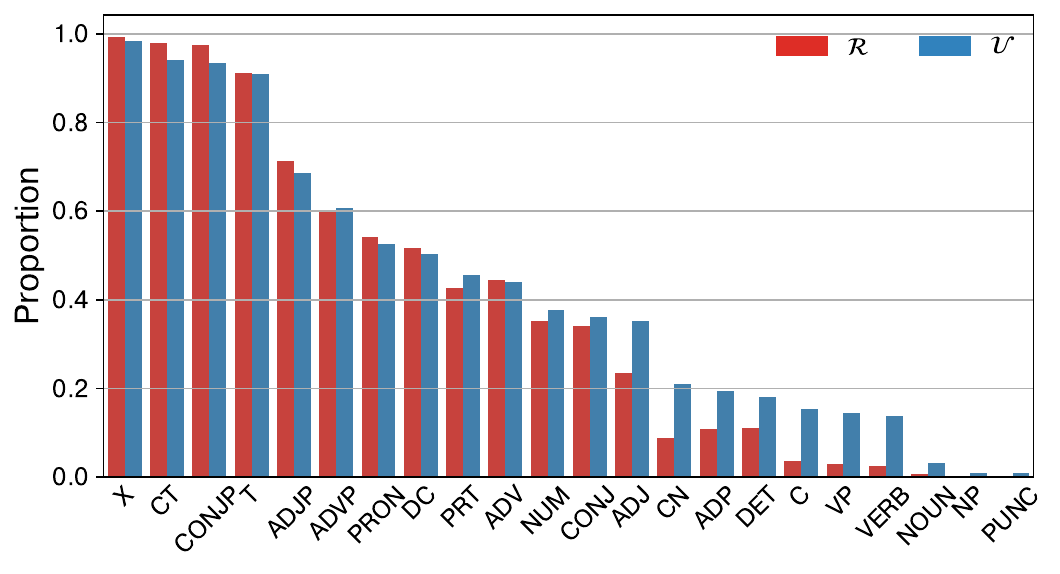}
    % \vspace{-10pt}
    \caption{Distribution of false positive predictions by \gpt for absent linguistic structures in input. All evaluated LLMs show very similar distribution}
\label{fig:over_conf}
\end{figure}

% In the previous sections, when evaluating the performance on specific linguistic structures, we skip the samples which contains no such type of structures. 
% \subsection{Prediction of Linguistic Structures}
We also observe that when a particular linguistic structure is absent in a given sample, LLMs still frequently make inaccurate predictions of its presence. Specifically for \gpt, we find that in 6,892 out of 21,000 queries (33.9\%), \gpt generates false positive predictions. Figure~\ref{fig:over_conf} shows the distribution of such errors across POS tag categories. The results show that \gpt often predict the existence of numerals (NUM), conjunctions (CONJ) and pronouns (PRON) when they are not present in the inputs. 
We conjecture that this behavior is due to biases in training data where certain words or structures co-occur frequently and the model learns to predict the presence of these words or structures based on relevant patterns in the training data, even when they don't exist in the input. For instance, if a sentence discusses quantities, the model might predict numerals. 
Therefore, false positive predictions for linguistic structures is common. 
In addition, all LLMs achieve higher recall than precision, especially all scales of LLaMA (see Table~\ref{tab:main_qp}), again indicating that LLMs tend to retrieve more false positives than false negatives. 

% In addition, the false positive predictions could be because LLMs generalize from their training data and apply these generalizations too broadly.

% We then investigate if over-confident predictions is correlated with linguistic characteristics of the input. Fig.~\ref{fig:overconf_vs_ind} illustrates the change of the number of over-confident predictions with respect to all 42 linguistic indices across all sentences in the corpus. Results show that overall \gpt tend to be over-confident when the input is short, easy and less rich inputs. For example, when there is only one vert in the sentence, \gpt is very likely to make a prediction, which can be a verb that do not exist in the input sentence, resulting in an over-confident prediction, i.e. 0 precision and recall. When the input sentence is complex with multiple verbs, \gpt acts more carefully. It either predicts a subset of ground truth verbs or makes no prediction, resulting in high precision and 0 precision, respectively.

\subsection{Model Capacity}
We observe that models with higher capacity show slightly better performance. We evaluate the effect of model capacity, measured by the number of parameters, in performing fine-grained linguistic annotation tasks by comparing two scales of Llama3, Llama2, and Mistral, see Table~\ref{tab:main_qp}.
% Figures~\ref{fig:model_scale_vicuna}~and~\ref{fig:model_scale_llama} in Appendix~\ref{sec:more_result}. 
All models show improved or maintained linguistic performance as their capacity increases. However, it's noteworthy that the performance advantage may not be significant enough compared to the increase of scale. Specifically, using a 10 times larger Llama3 and Llama2 only boosts F1 score by 5.8 and 5.0, and 7.3 and 5.5 on randomly and uniformly sampled data respectively. The performance gain is also smaller on uniformly sampled inputs across all LLMs, due to the diverse inputs with various linguistic complexity, which outweighs model scale. 
% , outperforms its other two versions on the POS tagging task under the SP strategy. This interesting result indicates that, while increased model capacity often leads to enhanced linguistic performance, it does not guarantee a better linguistic capability across all evaluation settings. 

\subsection{Dense model vs. Sparse model}
Scaling up LLMs with Mixture-of-Experts (MoE)~\citep{shazeer2017} in a sparse manner is a more efficient approach than dense scaling. We find that MoE can effectively boost LLM performance, see Mixtral 8x7b vs. Mistral-7b in Table~\ref{tab:main_qp}. The performance of the MoE-based model--Mixtral 8x7b--is also comparable to that of Llama2-70b, a dense model of similar scale. This suggests that sparsity in LLMs is not a key or limiting factor in their fine-grained linguistic annotation ability.

\section{Discussion}
\label{sec:disc}

\begin{table*}[t]
\small
\centering
\begin{tabular}{p{6em}|p{0.4\linewidth}|p{0.4\linewidth}}
\toprule
    Type & Expected output & Output by \gpt \\
    \midrule
    \multirow{2}{6em}{Easy example} & He-PRON remains-VERB chief-ADJ executive-NOUN officer-NOUN .-PUNC & He-PRON remains-VERB chief-ADJ \colorbox{lightred}{executive-ADJ} officer-ADV .-PUNC \\
    \midrule
    % Corrupted format & & \\
    % \cmidrule{2-9}
    % Spelling mistake & Quantitative & Quantitive \\
    % \cmidrule{2-9}
    % Hallucination & & \\
    % \cmidrule{2-9}
    \multirow{6}{*}{Confusion} & The-DET consensus-NOUN calls-VERB for-ADP a-DET 0.5-NUM \%-NOUN increase-NOUN in-ADP September-NOUN personal-ADJ income-NOUN and-CONJ a-DET 0.3-NUM \%-NOUN gain-NOUN in-ADP consumption-NOUN .-PUNC & The-PRON consensus-NOUN calls-VERB for-PRON a-PRON 0.5-ADJ \%-PUNC increase-VERB in-PRON September-NOUN personal-MISSING income-MISSING and-PRON a-PRON \colorbox{lightred}{0.3-ADJ} \%-PUNC \colorbox{lightred}{gain-VERB} in-PRON consumption-NOUN .-PUNC \\
    \midrule
    % Confusion & & \\
    % \cmidrule{2-9}
    % Distraction & & \\
    % \cmidrule{2-9}
    \multirow{6}{*}{Skip token} & The-DET department-NOUN has-VERB collected-VERB over-ADV \$-PUNC 6.5-NUM million-NUM from-ADP brokers-NOUN so-ADV far-ADV and-CONJ recommended-VERB more-ADJ than-ADP 30-NUM of-ADP them-PRON for-ADP criminal-ADJ prosecution-NOUN .-PUNC & \colorbox{lightred}{The} department-NOUN has-PRON collected-VERB over-PRON \$-PRON 6.5-PRON million-PRON \colorbox{lightred}{from} brokers-NOUN so-PRON far-PRON and-PRON recommended-VERB more-PRON than-PRON 30-NOUN of-PRON them-PRON for-PRON criminal-PRON prosecution-PRON .-PUNC \\
    % Inconsistency & IP & A fixed format across different samples & \textbf{Output for one sample}: ADV DET NOUN AUX VERB PRON ADJ SCONJ PRON AUX VERB PART VERB NOUN NOUN ADV ADP ADP VERB CONJ PRON ADP VERB ADV . \newline \textbf{Output for another sample}: It/PRON happened/VERB at/ADP Northrop/NOUN Corp./NOUN in/ADP Los/PROPN Angeles/PROPN ./PUNCT \\
    % \cmidrule{2-9}
    % Mixed tag set & IP & ADP ADJ ADJ NOUN PUNC NOUN NOUN VERB NUM PRT NUM & IN ADJ ADJ NOUN , NOUN NOUN VERB NUMBERS.prep. (\textbf{IN and , are from PTB tag set.}) \\
\bottomrule
  \end{tabular}
  \caption{Summary of inaccurate, low quality and erroneous linguistic content generated by \gpt.}
  \label{tab:mistake_example}
\end{table*}

\subsection{Limitations and Failure Cases}
\label{sec:error}
We identify significant limitations in the performance of the evaluated LLMs when responding to linguistic queries. Since \gpt is one of the most capable models with instruction-following capabilities in the evaluated LLMs, we mainly focus on its limitations.
% \paragraph{Failure on linguistically easy examples} 
\gpt may {\bf fail on linguistically easy examples}. For instance, it does not detect any of the 34 nouns that appear in the easiest linguistic examples in our dataset.
\gpt (and most evaluated LLMs) may occasionally {\bf skip tokens} in their responses. For example, they may skip tagging nouns or punctuation in inputs, which reduces their overall performance. 
They also {\bf generate ill-formatted outputs}, including missing tags or corrupted parse trees. This is unexpected given that \gpt have a good knowledge about the definition of the linguistic tasks and required format; see Appendix~\ref{sec:gpt_knowledge}, Figures~\ref{fig:gpt_pos}--\ref{fig:gpt_parsing}.
% \paragraph{Confusion} Sometimes \gpt confuses different POS tag sets (PTB vs. Universal), syntactic tags with POS tags. For \gpt, we encountered XX outputs tagged the PTB POS tag set, even if we explicitly ask it to use Universal POS tag set. 
% \paragraph{Inconsistency} 
% Despite the same prompt for every query, \gpt can vary its output considerably in terms of format. In the case of POS tagging, \gpt can format its output with a token, some random separator, a POS tag. Or it may output the sentence in a row and the corresponding POS tags in another row. This inconsistency hinders the predictability and stability of \gpt's output. 
%
% \paragraph{Mixed tag sets} \gpt may vary between Universal Dependency POS tag set and PennTreebank tag set across the samples. Sometimes, it even uses a mixture of different tag sets for a single sample.
% \paragraph{Easy to be distracted} 
% \gpt can be easily distracted by special characters appending to text. For example, ``single'' can be correctly recognized as an adjective but ``single-'' is not recognized. This highlights the instability and vulnerability of current LLMs and prompting methods.
\gpt (and most evaluated LLMs) may {\bf generate biased outputs}. 
We find that the evaluated LLMs are biased to output common tokens and concepts, such as nouns and pronouns, while neglecting uncommon ones. Table~\ref{tab:mistake_example} provides several example outputs. The first two show the tendency of \gpt to misclassify familiar structures such as nouns as adjectives or adverbs (first row) or verbs (second row). 
In addition, in may skip generating tags, see highlighted words ``The'' and ``from'' in Table 4. Note that the third example also shows \gpt mislabels many tokens, frequently replacing the correct POS tags with ``PRON'' or other incorrect tags.
% \flan do not seem to understand IP prompts, and the linguistic task and context in the instruction. 
% Though \gpt can sometimes detect it does not understand the instruction
% \footnote[1]{An example output is ``Unfortunately, I don't understand what you mean by \textit{citing the task}''.}
% , \flan does not have this capability.  
% \gpt sometimes converts the linguistic entity names to its abbreviations, such as returning ``adj'' when we asked ``Adjective'', and returning ``coord.conj'' when we asked `Coordinating Conjunction'. This may happen since the in pre-training data, these linguistic entities are usually refered to with their abbreviations. 

% \paragraph{Confusing information} It is also very common that \gpt mis-spells the linguistic entity names, even though they are present and highlighted in the prompt. For example, \gpt outputs ``Quantitive'' and ``Conjuction'', which are mis-spelled.

% \gpt sometimes fabricate linguistic entities that are not present in the prompt or do not exist. For example, \gpt may return ``Quantitative Phrase'', ``Quantitive Phrase``, and ``prepositionntclause''. Also, \gpt can mix up two entities and return ``Cardinal Phrase'', even though we only asked for ``Cardinal Number''. We present a subset of errors that \gpt made in Table~\ref{tab:error} due to limited space.

\subsection{Quality of Alignments}\label{sec:align}
The LLMs we consider for this study have instruction-following capabilities. However, their performances in following linguistic-related instructions vary considerably. We find that \gpt tends to follow instructions better than other LLMs evaluated in our experiments. % and few extra outputs.
On the other hand, \llama-2 and \llama-3 generate irrelevant outputs including auxiliary text and special characters such as ``\textbackslash'' and ``>'', even if we explicitly prompt them not to generate such characters. 
% For \vic and \flan, we observe a strong tendency of finding only a single output. \flan is especially inclined to generate descriptive and auxiliary text before the desired output, such as ``the noun in this sentence is:''
% , which incurred significant amount of effort for post-processing. We find it hard to consistently prevent this behaviour of \flan, through several trial of tweaking the prompt.
We note that the base model of \gpt has been fine-tuned on code data, which helps the model understand structured and instructional input. In addition, reinforcement learning from human feedback (RLHF)~\citep{NEURIPS2020_1f89885d} is a more effective method to align LLMs for following instructions than fine-tuning with conversation data~\citep{zheng2023judging} and instruction tuning~\citep{chung2022scaling}. In addition, \gpt has a much larger number of parameters, much larger capacity, than our other evaluated LLMs. 

We find that \gem rejects 24,630 prompts due to safety concerns related to sexual content, hate speech, harassment. Since we prompt \gem to perform linguistic tasks, we conclude \gem misinterpret linguistic queries with harmful contents, and it may have been superficially aligned~\citep{zhou2023lima} to restrict its ability when processing sensitive words in the prompt. It is also over-aligned in terms of security-related content, leading to degenerated and undesired behavior.\looseness-1

Furthermore, we find that small scales of \llama-2 and Mistral do not follow instructions. They sometimes simply echo back the input sentence without linguistic annotation, responding they don't understand what the task is, or ask for the input to be processed. We hypothesize that this is strongly correlated with the distribution of instruction-tuning data, where linguistic instructions do not appear frequently.

% \paragraph{Flexibility of prompt}
% Despite the high performance of \spt, it only works with tasks with structured output formats, which usually holds for low-level linguistic probing tasks (e.g. each token has a POS tag). However, for high-level tasks (e.g. readability score) or open-ended tasks, fixing a structured output format may not be possible, which significantly hinders the usability of \spt. Another disadvantage of \spt is that finding meaningful demonstrations may not be easy and may dramatically influence the final performance as discussed in many ICL work~\citep{pmlr-v139-zhao21c,chen2023relation}.

% \paragraph{Influence of system message}
% Many LLMs support providing a system message, which is pre-pended to the prompt to pre-define the behavior of the system. On \gpt, we do not observe a significant difference whether or not using system message. For all other evaluated models, we find using an empty system message leads to much better performance, which may be due to shorter prompt.

\subsection{Differences among Prompting Strategies} 
Prompting format and strategies differ in how they elicit knowledge from LLMs. However, we find that on identifying linguistic structures, adding in-context examples~\citep{NEURIPS2020_1457c0d6}, CoT~\citep{pmlr-v162-huang22a} or ReAct~\citep{yao2023react} provide only trivial performance gain over the plain prompt (0.05, 0.02 and 0.03 in F1 score respectively). We hypothesize that identifying linguistic structures, especially the complex ones, requires fundamental understanding of syntax and semantics, while CoT and ReAct focus on eliciting reasoning capabilities of LLMs, which is not sufficient. 
% On the other hand, the linguistic annotation performance depends more on input complexity.

\subsection{Potential Solutions}
% Since current LLMs are trained on web scale and diverse data, including considerable amount of linguistic data, and can generate coherent text, it is reasonable to hypothesize that these models encode linguistic knowledge in their parameter space. However, they may not closely follow linguistic instructions or be familiar with the task format, since the instruction-tuning datasets are general purpose and do not contain many linguistic-related samples. Thus, we argue that a potential solution to make LLMs better under such linguistic examinations is to create linguistic instruction-tuning samples and fine-tuning approaches.
Addressing the above limitations and biases requires developing effective data curation and training strategies using a linguist-in-the-loop process. Linguistically equitable and diverse datasets with balanced presence of linguistic structures (that specifically avoid overrepresentation of linguistically easy samples) are essential for NLP and for analyzing and understanding LLMs from a linguistic perspective. In what follows, we provide several avenues for investigating the above limitations. 

\paragraph{Direct Training:} Fine-tuning LLMs with targeted challenging examples, like those carrying complex sentence structures, or augmenting data to increase exposure to challenging examples can improve LLM's performance on fine-grained linguistic annotation tasks~\citep{nguyen-etal-2024-multi}. The resulting computational costs can be alleviated through Parameter-Efficient Fine-Tuning techniques~\citep{hu2022lora,su-etal-2023-exploring}.

\paragraph{Better Instructions:} Designing linguistic instructions with sufficient context information to improve contextual understanding can potentially guide the model in handling complex structures. However, it would be challenging to generalize instructions to all linguistic structures and LLMs.

\paragraph{Curriculum Learning:} LLM's performance on challenging linguistic structures could be improved by gradually training through a linguistic curriculum~\citep{elgaar-amiri-2023-hucurl}. A curriculum is a planned sequence of learning materials (a training paradigm) and an effective one can make learning efficient and effective for humans~\citep{Nishimura2018-en,Tabibian2019-ic} and computers~\citep{10.1145/1553374.1553380}. 
Curriculum learning techniques can present progressively increase the complexity of the linguistic structure of training samples, e.g. starting with easier structures before more complex ones to potentially improve LLM's performance on fine-grained linguistic annotation tasks. 

\paragraph{Retrieval Augmented Generation:} Incorporating documents with relevant linguistic knowledge retrieved from trustworthy sources can complement LLMs' knowledge~\citep{NEURIPS2020_6b493230}. For example, definitions of complex syntactic structures such as clauses and T-units can be retrieved to support more accurate analysis and generation. However, care must be taken to mitigate potential biases introduced within retrieval models~\citep{ziems-etal-2024-measuring,cheng-etal-2025-equalizeir}. 

\paragraph{Tool Learning:} LLMs can be trained to use tools~\citep{schick2023toolformer}, either by updating their parametric knowledge or interacting with tools directly. Training LLMs to use external linguistic tools, such as those discussed in this work~\citep{Lu2010,Lu2012,lee-etal-2021-pushing,lee-lee-2023-lftk}, can potentially improve LLMs’ capabilities on fine-grained linguistic tasks by complementing their internal representations with structured linguistic knowledge.

\paragraph{Human-in-the-Loop:} Using a linguist-in-the-loop approach can provide a valuable feedback for refining model outputs. Expert input can help correct linguistic errors, mitigate biases, and guide the model toward more accurate and interpretable language understanding~\citep{parrish-etal-2021-putting-linguist}.
\section{Conclusion}
We empirically study the ability of recent LLMs in annotating linguistic structures at different levels of linguistic complexity. 
Our study determines 
how accurately recent LLMs can detect complex linguistic structures in input text, 
which linguistic structures represent the blind spots of recent LLMs (the most challenging for LLMs), and 
how the performance of LLMs varies across different levels of linguistic complexity of inputs.
Our findings show a tendency to overestimate the linguistic capabilities of LLMs in previous research, which mainly stems from the prevalence of linguistically easy examples in NLP datasets. To address this gap, we uniformly sample data from different linguistic complexity groups, to improve the reliability of evaluating LLMs' performance. 
Among all evaluated LLMs, Llama3-70b, Llama3-8b, and \gpt show relatively better performance in responding to linguistic queries--though overall performance remains low. We outline several potential solutions to address these limitations. 
\section*{Limitations}
% Our work has certain limitations. First, we only perform the study Penn Treebank corpus and on a limited set of LLMs.

% The performance of LLMs is highly dependent on the quality and design of the prompts used. Biased or poorly crafted prompts can produce skewed results. 
Although we carefully developed and experimented with different prompting strategies, prompting cannot fully replace methods that directly analyze model’s probability distributions over outputs~\citep{hu-levy-2023-prompting,kuribayashi-etal-2024-psychometric}. In addition, we did not investigate the ability of LLMs on a wider range of linguistic queries. For examples, linguistic structures related to \emph{discourse complexity}~\citep{feng-etal-2010-comparison,guinaudeau-strube-2013-graph,bedi2015automated}, which determines the complexity of higher-level structures and flow of language beyond individual phrases or sentences, need to investigated. 
Finally, understanding why a closed-source LLM produces a specific output can be challenging. This is a key challenge for deeper understanding of LLMs through theoretically-motivated linguistic probing techniques~\citep{linzen-etal-2016-assessing,warstadt-etal-2020-blimp-benchmark,hu-etal-2020-ocnli}, and limits our ability in providing insights into their potential weaknesses.

% \section*{Broader Impact and Ethical Statement}
% Our findings help researchers identify areas for improving LLMs and have the potential to inspire the development of new methods to address LLMs's limitations in fine-grained linguistic annotation. In addition, our work reveals significant disparity in linguistic complexity of data in well-established NLP benchmarks, which indicates potential overestimation of the capabilities of recent LLMs in existing research.
% We acknowledge and emphasize the ethical mindfulness throughout the design, training, and applying the models investigated in this work. 

% \section*{Ethical Considerations}

\bibliography{anthology,custom}

\begin{thebibliography}{74}
\expandafter\ifx\csname natexlab\endcsname\relax\def\natexlab#1{#1}\fi

\bibitem[{Alajrami and Aletras(2022)}]{alajrami-aletras-2022-pre}
Ahmed Alajrami and Nikolaos Aletras. 2022.
\newblock \href {https://doi.org/10.18653/v1/2022.acl-short.16} {How does the
  pre-training objective affect what large language models learn about
  linguistic properties?}
\newblock In \emph{Proceedings of the 60th Annual Meeting of the Association
  for Computational Linguistics (Volume 2: Short Papers)}, pages 131--147,
  Dublin, Ireland. Association for Computational Linguistics.

\bibitem[{Bedi et~al.(2015)Bedi, Carrillo, Cecchi, Slezak, Sigman, Mota,
  Ribeiro, Javitt, Copelli, and Corcoran}]{bedi2015automated}
Gillinder Bedi, Facundo Carrillo, Guillermo~A Cecchi, Diego~Fern{\'a}ndez
  Slezak, Mariano Sigman, Nat{\'a}lia~B Mota, Sidarta Ribeiro, Daniel~C Javitt,
  Mauro Copelli, and Cheryl~M Corcoran. 2015.
\newblock \href {https://pubmed.ncbi.nlm.nih.gov/27336038/} {Automated analysis
  of free speech predicts psychosis onset in high-risk youths}.
\newblock \emph{npj Schizophrenia}, 1(1):1--7.

\bibitem[{Bengio et~al.(2009)Bengio, Louradour, Collobert, and
  Weston}]{10.1145/1553374.1553380}
Yoshua Bengio, J\'{e}r\^{o}me Louradour, Ronan Collobert, and Jason Weston.
  2009.
\newblock \href {https://doi.org/10.1145/1553374.1553380} {Curriculum
  learning}.
\newblock In \emph{Proceedings of the 26th Annual International Conference on
  Machine Learning}, ICML '09, page 41–48, New York, NY, USA. Association for
  Computing Machinery.

\bibitem[{Biber et~al.(2020)Biber, Gray, Staples, and
  Egbert}]{biber2020investigating}
Douglas Biber, Bethany Gray, Shelley Staples, and Jesse Egbert. 2020.
\newblock \href
  {https://www.sciencedirect.com/science/article/abs/pii/S1475158519305909}
  {Investigating grammatical complexity in l2 english writing research:
  Linguistic description versus predictive measurement}.
\newblock \emph{Journal of English for Academic Purposes}, 46:100869.

\bibitem[{Blevins et~al.(2023)Blevins, Gonen, and
  Zettlemoyer}]{blevins-etal-2023-prompting}
Terra Blevins, Hila Gonen, and Luke Zettlemoyer. 2023.
\newblock \href {https://doi.org/10.18653/v1/2023.acl-long.367} {Prompting
  language models for linguistic structure}.
\newblock In \emph{Proceedings of the 61st Annual Meeting of the Association
  for Computational Linguistics (Volume 1: Long Papers)}, pages 6649--6663,
  Toronto, Canada. Association for Computational Linguistics.

\bibitem[{Brown et~al.(2020)Brown, Mann, Ryder, Subbiah, Kaplan, Dhariwal,
  Neelakantan, Shyam, Sastry, Askell, Agarwal, Herbert-Voss, Krueger, Henighan,
  Child, Ramesh, Ziegler, Wu, Winter, Hesse, Chen, Sigler, Litwin, Gray, Chess,
  Clark, Berner, McCandlish, Radford, Sutskever, and
  Amodei}]{NEURIPS2020_1457c0d6}
Tom Brown, Benjamin Mann, Nick Ryder, Melanie Subbiah, Jared~D Kaplan, Prafulla
  Dhariwal, Arvind Neelakantan, Pranav Shyam, Girish Sastry, Amanda Askell,
  Sandhini Agarwal, Ariel Herbert-Voss, Gretchen Krueger, Tom Henighan, Rewon
  Child, Aditya Ramesh, Daniel Ziegler, Jeffrey Wu, Clemens Winter, Chris
  Hesse, Mark Chen, Eric Sigler, Mateusz Litwin, Scott Gray, Benjamin Chess,
  Jack Clark, Christopher Berner, Sam McCandlish, Alec Radford, Ilya Sutskever,
  and Dario Amodei. 2020.
\newblock \href
  {https://proceedings.neurips.cc/paper_files/paper/2020/file/1457c0d6bfcb4967418bfb8ac142f64a-Paper.pdf}
  {Language models are few-shot learners}.
\newblock In \emph{Advances in Neural Information Processing Systems},
  volume~33, pages 1877--1901. Curran Associates, Inc.

\bibitem[{Chen et~al.(2024)Chen, Shwartz-Ziv, Cho, Leavitt, and
  Saphra}]{chen2024sudden}
Angelica Chen, Ravid Shwartz-Ziv, Kyunghyun Cho, Matthew~L Leavitt, and Naomi
  Saphra. 2024.
\newblock \href {https://openreview.net/forum?id=MO5PiKHELW} {Sudden drops in
  the loss: Syntax acquisition, phase transitions, and simplicity bias in
  {MLM}s}.
\newblock In \emph{The Twelfth International Conference on Learning
  Representations}.

\bibitem[{Cheng and Amiri(2025)}]{cheng-etal-2025-equalizeir}
Jiali Cheng and Hadi Amiri. 2025.
\newblock Equalizeir: Mitigating linguistic biases in retrieval models.
\newblock In \emph{Proceedings of the 2025 Conference of the North American
  Chapter of the Association for Computational Linguistics}. Association for
  Computational Linguistics.

\bibitem[{Chung et~al.(2022)Chung, Hou, Longpre, Zoph, Tay, Fedus, Li, Wang,
  Dehghani, Brahma, Webson, Gu, Dai, Suzgun, Chen, Chowdhery, Castro-Ros,
  Pellat, Robinson, Valter, Narang, Mishra, Yu, Zhao, Huang, Dai, Yu, Petrov,
  Chi, Dean, Devlin, Roberts, Zhou, Le, and Wei}]{chung2022scaling}
Hyung~Won Chung, Le~Hou, Shayne Longpre, Barret Zoph, Yi~Tay, William Fedus,
  Yunxuan Li, Xuezhi Wang, Mostafa Dehghani, Siddhartha Brahma, Albert Webson,
  Shixiang~Shane Gu, Zhuyun Dai, Mirac Suzgun, Xinyun Chen, Aakanksha
  Chowdhery, Alex Castro-Ros, Marie Pellat, Kevin Robinson, Dasha Valter,
  Sharan Narang, Gaurav Mishra, Adams Yu, Vincent Zhao, Yanping Huang, Andrew
  Dai, Hongkun Yu, Slav Petrov, Ed~H. Chi, Jeff Dean, Jacob Devlin, Adam
  Roberts, Denny Zhou, Quoc~V. Le, and Jason Wei. 2022.
\newblock \href {http://arxiv.org/abs/2210.11416} {Scaling
  instruction-finetuned language models}.

\bibitem[{Clark et~al.(2019)Clark, Khandelwal, Levy, and
  Manning}]{clark-etal-2019-bert}
Kevin Clark, Urvashi Khandelwal, Omer Levy, and Christopher~D. Manning. 2019.
\newblock \href {https://doi.org/10.18653/v1/W19-4828} {What does {BERT} look
  at? an analysis of {BERT}`s attention}.
\newblock In \emph{Proceedings of the 2019 ACL Workshop BlackboxNLP: Analyzing
  and Interpreting Neural Networks for NLP}, pages 276--286, Florence, Italy.
  Association for Computational Linguistics.

\bibitem[{Devlin et~al.(2019)Devlin, Chang, Lee, and
  Toutanova}]{devlin-etal-2019-bert}
Jacob Devlin, Ming-Wei Chang, Kenton Lee, and Kristina Toutanova. 2019.
\newblock \href {https://doi.org/10.18653/v1/N19-1423} {{BERT}: Pre-training of
  deep bidirectional transformers for language understanding}.
\newblock In \emph{Proceedings of the 2019 Conference of the North {A}merican
  Chapter of the Association for Computational Linguistics: Human Language
  Technologies, Volume 1 (Long and Short Papers)}, pages 4171--4186,
  Minneapolis, Minnesota. Association for Computational Linguistics.

\bibitem[{Durrani et~al.(2020)Durrani, Sajjad, Dalvi, and
  Belinkov}]{durrani-etal-2020-analyzing}
Nadir Durrani, Hassan Sajjad, Fahim Dalvi, and Yonatan Belinkov. 2020.
\newblock \href {https://doi.org/10.18653/v1/2020.emnlp-main.395} {Analyzing
  individual neurons in pre-trained language models}.
\newblock In \emph{Proceedings of the 2020 Conference on Empirical Methods in
  Natural Language Processing (EMNLP)}, pages 4865--4880, Online. Association
  for Computational Linguistics.

\bibitem[{Elgaar and Amiri(2023{\natexlab{a}})}]{elgaar-amiri-2023-hucurl}
Mohamed Elgaar and Hadi Amiri. 2023{\natexlab{a}}.
\newblock \href {https://doi.org/10.18653/v1/2023.acl-long.104} {{H}u{C}url:
  Human-induced curriculum discovery}.
\newblock In \emph{Proceedings of the 61st Annual Meeting of the Association
  for Computational Linguistics (Volume 1: Long Papers)}, pages 1862--1877,
  Toronto, Canada. Association for Computational Linguistics.

\bibitem[{Elgaar and Amiri(2023{\natexlab{b}})}]{elgaar-amiri-2023-ling}
Mohamed Elgaar and Hadi Amiri. 2023{\natexlab{b}}.
\newblock \href {https://doi.org/10.18653/v1/2023.emnlp-main.834} {Ling-{CL}:
  Understanding {NLP} models through linguistic curricula}.
\newblock In \emph{Proceedings of the 2023 Conference on Empirical Methods in
  Natural Language Processing}, pages 13526--13542, Singapore. Association for
  Computational Linguistics.

\bibitem[{Ettinger(2020)}]{ettinger-2020-bert}
Allyson Ettinger. 2020.
\newblock \href {https://doi.org/10.1162/tacl_a_00298} {What {BERT} is not:
  Lessons from a new suite of psycholinguistic diagnostics for language
  models}.
\newblock \emph{Transactions of the Association for Computational Linguistics},
  8:34--48.

\bibitem[{Feng et~al.(2009)Feng, Elhadad, and
  Huenerfauth}]{feng-etal-2009-cognitively}
Lijun Feng, No{\'e}mie Elhadad, and Matt Huenerfauth. 2009.
\newblock \href {https://aclanthology.org/E09-1027/} {Cognitively motivated
  features for readability assessment}.
\newblock In \emph{Proceedings of the 12th Conference of the {E}uropean Chapter
  of the {ACL} ({EACL} 2009)}, pages 229--237, Athens, Greece. Association for
  Computational Linguistics.

\bibitem[{Feng et~al.(2010)Feng, Jansche, Huenerfauth, and
  Elhadad}]{feng-etal-2010-comparison}
Lijun Feng, Martin Jansche, Matt Huenerfauth, and No{\'e}mie Elhadad. 2010.
\newblock \href {https://aclanthology.org/C10-2032/} {A comparison of features
  for automatic readability assessment}.
\newblock In \emph{Coling 2010: Posters}, pages 276--284, Beijing, China.
  Coling 2010 Organizing Committee.

\bibitem[{Guinaudeau and Strube(2013)}]{guinaudeau-strube-2013-graph}
Camille Guinaudeau and Michael Strube. 2013.
\newblock \href {https://aclanthology.org/P13-1010/} {Graph-based local
  coherence modeling}.
\newblock In \emph{Proceedings of the 51st Annual Meeting of the Association
  for Computational Linguistics (Volume 1: Long Papers)}, pages 93--103, Sofia,
  Bulgaria. Association for Computational Linguistics.

\bibitem[{Housen et~al.(2019)Housen, De~Clercq, Kuiken, and
  Vedder}]{housen2019multiple}
Alex Housen, Bastien De~Clercq, Folkert Kuiken, and Ineke Vedder. 2019.
\newblock \href
  {https://journals.sagepub.com/doi/full/10.1177/0267658318809765} {Multiple
  approaches to complexity in second language research}.
\newblock \emph{Second language research}, 35(1):3--21.

\bibitem[{Hu et~al.(2022)Hu, yelong shen, Wallis, Allen-Zhu, Li, Wang, Wang,
  and Chen}]{hu2022lora}
Edward~J Hu, yelong shen, Phillip Wallis, Zeyuan Allen-Zhu, Yuanzhi Li, Shean
  Wang, Lu~Wang, and Weizhu Chen. 2022.
\newblock \href {https://openreview.net/forum?id=nZeVKeeFYf9} {Lo{RA}: Low-rank
  adaptation of large language models}.
\newblock In \emph{International Conference on Learning Representations}.

\bibitem[{Hu et~al.(2020)Hu, Richardson, Xu, Li, K{\"u}bler, and
  Moss}]{hu-etal-2020-ocnli}
Hai Hu, Kyle Richardson, Liang Xu, Lu~Li, Sandra K{\"u}bler, and Lawrence Moss.
  2020.
\newblock \href {https://doi.org/10.18653/v1/2020.findings-emnlp.314} {{OCNLI}:
  {O}riginal {C}hinese {N}atural {L}anguage {I}nference}.
\newblock In \emph{Findings of the Association for Computational Linguistics:
  EMNLP 2020}, pages 3512--3526, Online. Association for Computational
  Linguistics.

\bibitem[{Hu and Levy(2023)}]{hu-levy-2023-prompting}
Jennifer Hu and Roger Levy. 2023.
\newblock \href {https://doi.org/10.18653/v1/2023.emnlp-main.306} {Prompting is
  not a substitute for probability measurements in large language models}.
\newblock In \emph{Proceedings of the 2023 Conference on Empirical Methods in
  Natural Language Processing}, pages 5040--5060, Singapore. Association for
  Computational Linguistics.

\bibitem[{Huang et~al.(2022)Huang, Abbeel, Pathak, and
  Mordatch}]{pmlr-v162-huang22a}
Wenlong Huang, Pieter Abbeel, Deepak Pathak, and Igor Mordatch. 2022.
\newblock \href {https://proceedings.mlr.press/v162/huang22a.html} {Language
  models as zero-shot planners: Extracting actionable knowledge for embodied
  agents}.
\newblock In \emph{Proceedings of the 39th International Conference on Machine
  Learning}.

\bibitem[{Huebner et~al.(2021)Huebner, Sulem, Cynthia, and
  Roth}]{huebner-etal-2021-babyberta}
Philip~A. Huebner, Elior Sulem, Fisher Cynthia, and Dan Roth. 2021.
\newblock \href {https://doi.org/10.18653/v1/2021.conll-1.49} {{B}aby{BERT}a:
  Learning more grammar with small-scale child-directed language}.
\newblock In \emph{Proceedings of the 25th Conference on Computational Natural
  Language Learning}, pages 624--646, Online. Association for Computational
  Linguistics.

\bibitem[{Jiang et~al.(2023)Jiang, Sablayrolles, Mensch, Bamford, Chaplot,
  Casas, Bressand, Lengyel, Lample, Saulnier et~al.}]{jiang2023mistral}
Albert~Q Jiang, Alexandre Sablayrolles, Arthur Mensch, Chris Bamford,
  Devendra~Singh Chaplot, Diego de~las Casas, Florian Bressand, Gianna Lengyel,
  Guillaume Lample, Lucile Saulnier, et~al. 2023.
\newblock \href {https://arxiv.org/abs/2310.06825} {Mistral 7b}.
\newblock \emph{arXiv preprint arXiv:2310.06825}.

\bibitem[{Jiang et~al.(2024)Jiang, Sablayrolles, Roux, Mensch, Savary, Bamford,
  Chaplot, Casas, Hanna, Bressand et~al.}]{jiang2024mixtral}
Albert~Q Jiang, Alexandre Sablayrolles, Antoine Roux, Arthur Mensch, Blanche
  Savary, Chris Bamford, Devendra~Singh Chaplot, Diego de~las Casas, Emma~Bou
  Hanna, Florian Bressand, et~al. 2024.
\newblock \href {https://arxiv.org/abs/2401.04088} {Mixtral of experts}.
\newblock \emph{arXiv preprint arXiv:2401.04088}.

\bibitem[{Kuribayashi et~al.(2024)Kuribayashi, Oseki, and
  Baldwin}]{kuribayashi-etal-2024-psychometric}
Tatsuki Kuribayashi, Yohei Oseki, and Timothy Baldwin. 2024.
\newblock \href {https://doi.org/10.18653/v1/2024.findings-naacl.129}
  {Psychometric predictive power of large language models}.
\newblock In \emph{Findings of the Association for Computational Linguistics:
  NAACL 2024}, pages 1983--2005, Mexico City, Mexico. Association for
  Computational Linguistics.

\bibitem[{Lee et~al.(2021)Lee, Jang, and Lee}]{lee-etal-2021-pushing}
Bruce~W. Lee, Yoo~Sung Jang, and Jason Lee. 2021.
\newblock \href {https://doi.org/10.18653/v1/2021.emnlp-main.834} {Pushing on
  text readability assessment: A transformer meets handcrafted linguistic
  features}.
\newblock In \emph{Proceedings of the 2021 Conference on Empirical Methods in
  Natural Language Processing}, pages 10669--10686, Online and Punta Cana,
  Dominican Republic. Association for Computational Linguistics.

\bibitem[{Lee and Lee(2023)}]{lee-lee-2023-lftk}
Bruce~W. Lee and Jason Lee. 2023.
\newblock \href {https://doi.org/10.18653/v1/2023.bea-1.1} {{LFTK}: Handcrafted
  features in computational linguistics}.
\newblock In \emph{Proceedings of the 18th Workshop on Innovative Use of NLP
  for Building Educational Applications (BEA 2023)}, pages 1--19, Toronto,
  Canada. Association for Computational Linguistics.

\bibitem[{Lewis et~al.(2020)Lewis, Perez, Piktus, Petroni, Karpukhin, Goyal,
  K\"{u}ttler, Lewis, Yih, Rockt\"{a}schel, Riedel, and
  Kiela}]{NEURIPS2020_6b493230}
Patrick Lewis, Ethan Perez, Aleksandra Piktus, Fabio Petroni, Vladimir
  Karpukhin, Naman Goyal, Heinrich K\"{u}ttler, Mike Lewis, Wen-tau Yih, Tim
  Rockt\"{a}schel, Sebastian Riedel, and Douwe Kiela. 2020.
\newblock \href
  {https://proceedings.neurips.cc/paper_files/paper/2020/file/6b493230205f780e1bc26945df7481e5-Paper.pdf}
  {Retrieval-augmented generation for knowledge-intensive nlp tasks}.
\newblock In \emph{Advances in Neural Information Processing Systems},
  volume~33, pages 9459--9474. Curran Associates, Inc.

\bibitem[{Li et~al.(2024)Li, Su, Huang, Cheng, Hu, Zhang, Wang, Qin, Wang,
  Lindquist, Liu, and Zhang}]{li2024emotion}
Ming Li, Yusheng Su, Hsiu-Yuan Huang, Jiali Cheng, Xin Hu, Xinmiao Zhang,
  Huadong Wang, Yujia Qin, Xiaozhi Wang, Kristen~A. Lindquist, Zhiyuan Liu, and
  Dan Zhang. 2024.
\newblock \href {https://doi.org/10.1016/j.isci.2024.111401} {Language-specific
  representation of emotion-concept knowledge causally supports emotion
  inference}.
\newblock \emph{iScience}, 27(12).

\bibitem[{Linzen et~al.(2016)Linzen, Dupoux, and
  Goldberg}]{linzen-etal-2016-assessing}
Tal Linzen, Emmanuel Dupoux, and Yoav Goldberg. 2016.
\newblock \href {https://doi.org/10.1162/tacl_a_00115} {Assessing the ability
  of {LSTM}s to learn syntax-sensitive dependencies}.
\newblock \emph{Transactions of the Association for Computational Linguistics},
  4:521--535.

\bibitem[{Lu(2010)}]{Lu2010}
Xiaofei Lu. 2010.
\newblock \href
  {https://www.jbe-platform.com/docserver/fulltext/ijcl.15.4.02lu.pdf}
  {Automatic analysis of syntactic complexity in second language writing}.
\newblock \emph{International journal of corpus linguistics}, 15(4):474--496.

\bibitem[{Lu(2012)}]{Lu2012}
Xiaofei Lu. 2012.
\newblock \href
  {https://onlinelibrary.wiley.com/doi/full/10.1111/j.1540-4781.2011.01232_1.x}
  {The relationship of lexical richness to the quality of {ESL} learners' oral
  narratives}.
\newblock \emph{Mod. Lang. J.}, 96(2):190--208.

\bibitem[{Malvern et~al.(2004)Malvern, Richards, Chipere, and
  Dur{\'a}n}]{malvern2004lexical}
David Malvern, Brian Richards, Ngoni Chipere, and Pilar Dur{\'a}n. 2004.
\newblock \href {https://link.springer.com/book/10.1057/9780230511804}
  {\emph{Lexical diversity and language development}}.
\newblock Springer.

\bibitem[{Manning et~al.(2014)Manning, Surdeanu, Bauer, Finkel, Bethard, and
  McClosky}]{manning-etal-2014-stanford}
Christopher Manning, Mihai Surdeanu, John Bauer, Jenny Finkel, Steven Bethard,
  and David McClosky. 2014.
\newblock \href {https://doi.org/10.3115/v1/P14-5010} {The {S}tanford
  {C}ore{NLP} natural language processing toolkit}.
\newblock In \emph{Proceedings of 52nd Annual Meeting of the Association for
  Computational Linguistics: System Demonstrations}, pages 55--60, Baltimore,
  Maryland. Association for Computational Linguistics.

\bibitem[{Marcus et~al.(1993)Marcus, Santorini, and
  Marcinkiewicz}]{marcus-etal-1993-building}
Mitchell~P. Marcus, Beatrice Santorini, and Mary~Ann Marcinkiewicz. 1993.
\newblock \href {https://aclanthology.org/J93-2004/} {Building a large
  annotated corpus of {E}nglish: The {P}enn {T}reebank}.
\newblock \emph{Computational Linguistics}, 19(2):313--330.

\bibitem[{Nguyen et~al.(2024)Nguyen, Chen, and Zhou}]{nguyen-etal-2024-multi}
Dang Nguyen, Jiuhai Chen, and Tianyi Zhou. 2024.
\newblock \href {https://doi.org/10.18653/v1/2024.findings-acl.257}
  {Multi-objective linguistic control of large language models}.
\newblock In \emph{Findings of the Association for Computational Linguistics:
  ACL 2024}, pages 4336--4347, Bangkok, Thailand. Association for Computational
  Linguistics.

\bibitem[{Nishimura(2018)}]{Nishimura2018-en}
Joel Nishimura. 2018.
\newblock \href {https://arxiv.org/abs/1805.00051} {Critically slow learning in
  flashcard learning models}.
\newblock \emph{Chaos}, 28(8):083115.

\bibitem[{Ortega(2003)}]{ortega2003syntactic}
Lourdes Ortega. 2003.
\newblock \href
  {https://academic.oup.com/applij/article-abstract/24/4/492/213639?redirectedFrom=fulltext}
  {Syntactic complexity measures and their relationship to l2 proficiency: A
  research synthesis of college-level l2 writing}.
\newblock \emph{Applied linguistics}, 24(4):492--518.

\bibitem[{Ouyang et~al.(2022)Ouyang, Wu, Jiang, Almeida, Wainwright, Mishkin,
  Zhang, Agarwal, Slama, Gray, Schulman, Hilton, Kelton, Miller, Simens,
  Askell, Welinder, Christiano, Leike, and Lowe}]{ouyang2022training}
Long Ouyang, Jeffrey Wu, Xu~Jiang, Diogo Almeida, Carroll Wainwright, Pamela
  Mishkin, Chong Zhang, Sandhini Agarwal, Katarina Slama, Alex Gray, John
  Schulman, Jacob Hilton, Fraser Kelton, Luke Miller, Maddie Simens, Amanda
  Askell, Peter Welinder, Paul Christiano, Jan Leike, and Ryan Lowe. 2022.
\newblock \href {https://openreview.net/forum?id=TG8KACxEON} {Training language
  models to follow instructions with human feedback}.
\newblock In \emph{Advances in Neural Information Processing Systems}.

\bibitem[{Parrish et~al.(2021)Parrish, Huang, Agha, Lee, Nangia, Warstadt,
  Aggarwal, Allaway, Linzen, and Bowman}]{parrish-etal-2021-putting-linguist}
Alicia Parrish, William Huang, Omar Agha, Soo-Hwan Lee, Nikita Nangia, Alexia
  Warstadt, Karmanya Aggarwal, Emily Allaway, Tal Linzen, and Samuel~R. Bowman.
  2021.
\newblock \href {https://doi.org/10.18653/v1/2021.findings-emnlp.421} {Does
  putting a linguist in the loop improve {NLU} data collection?}
\newblock In \emph{Findings of the Association for Computational Linguistics:
  EMNLP 2021}, pages 4886--4901, Punta Cana, Dominican Republic. Association
  for Computational Linguistics.

\bibitem[{Schick et~al.(2023)Schick, Dwivedi-Yu, Dessi, Raileanu, Lomeli,
  Hambro, Zettlemoyer, Cancedda, and Scialom}]{schick2023toolformer}
Timo Schick, Jane Dwivedi-Yu, Roberto Dessi, Roberta Raileanu, Maria Lomeli,
  Eric Hambro, Luke Zettlemoyer, Nicola Cancedda, and Thomas Scialom. 2023.
\newblock \href {https://openreview.net/forum?id=Yacmpz84TH} {Toolformer:
  Language models can teach themselves to use tools}.
\newblock In \emph{Thirty-seventh Conference on Neural Information Processing
  Systems}.

\bibitem[{Sharma et~al.(2023)Sharma, Muralidhar, and
  Ramakrishnan}]{sharma-etal-2023-learning}
Mandar Sharma, Nikhil Muralidhar, and Naren Ramakrishnan. 2023.
\newblock \href {https://doi.org/10.18653/v1/2023.acl-long.340} {Learning
  non-linguistic skills without sacrificing linguistic proficiency}.
\newblock In \emph{Proceedings of the 61st Annual Meeting of the Association
  for Computational Linguistics (Volume 1: Long Papers)}, pages 6178--6191,
  Toronto, Canada. Association for Computational Linguistics.

\bibitem[{Shazeer et~al.(2017)Shazeer, Mirhoseini, Maziarz, Davis, Le, Hinton,
  and Dean}]{shazeer2017}
Noam Shazeer, *Azalia Mirhoseini, *Krzysztof Maziarz, Andy Davis, Quoc Le,
  Geoffrey Hinton, and Jeff Dean. 2017.
\newblock \href {https://openreview.net/forum?id=B1ckMDqlg} {Outrageously large
  neural networks: The sparsely-gated mixture-of-experts layer}.
\newblock In \emph{International Conference on Learning Representations}.

\bibitem[{Shen et~al.(2021)Shen, Yao, Kiela, Keutzer, and
  Mahoney}]{shen-etal-2021-whats}
Sheng Shen, Zhewei Yao, Douwe Kiela, Kurt Keutzer, and Michael Mahoney. 2021.
\newblock \href {https://doi.org/10.18653/v1/2021.emnlp-main.231} {What`s
  hidden in a one-layer randomly weighted transformer?}
\newblock In \emph{Proceedings of the 2021 Conference on Empirical Methods in
  Natural Language Processing}, pages 2914--2921, Online and Punta Cana,
  Dominican Republic. Association for Computational Linguistics.

\bibitem[{Shen et~al.(2018)Shen, Lin, Jacob, Sordoni, Courville, and
  Bengio}]{shen-etal-2018-straight}
Yikang Shen, Zhouhan Lin, Athul~Paul Jacob, Alessandro Sordoni, Aaron
  Courville, and Yoshua Bengio. 2018.
\newblock \href {https://doi.org/10.18653/v1/P18-1108} {Straight to the tree:
  Constituency parsing with neural syntactic distance}.
\newblock In \emph{Proceedings of the 56th Annual Meeting of the Association
  for Computational Linguistics (Volume 1: Long Papers)}, pages 1171--1180,
  Melbourne, Australia. Association for Computational Linguistics.

\bibitem[{Sinha et~al.(2021)Sinha, Parthasarathi, Pineau, and
  Williams}]{sinha-etal-2021-unnatural}
Koustuv Sinha, Prasanna Parthasarathi, Joelle Pineau, and Adina Williams. 2021.
\newblock \href {https://doi.org/10.18653/v1/2021.acl-long.569} {{UnNatural}
  {L}anguage {I}nference}.
\newblock In \emph{Proceedings of the 59th Annual Meeting of the Association
  for Computational Linguistics and the 11th International Joint Conference on
  Natural Language Processing (Volume 1: Long Papers)}, pages 7329--7346,
  Online. Association for Computational Linguistics.

\bibitem[{Stiennon et~al.(2020)Stiennon, Ouyang, Wu, Ziegler, Lowe, Voss,
  Radford, Amodei, and Christiano}]{NEURIPS2020_1f89885d}
Nisan Stiennon, Long Ouyang, Jeffrey Wu, Daniel Ziegler, Ryan Lowe, Chelsea
  Voss, Alec Radford, Dario Amodei, and Paul~F Christiano. 2020.
\newblock \href
  {https://proceedings.neurips.cc/paper_files/paper/2020/file/1f89885d556929e98d3ef9b86448f951-Paper.pdf}
  {Learning to summarize with human feedback}.
\newblock In \emph{Advances in Neural Information Processing Systems}.

\bibitem[{Su et~al.(2023)Su, Chan, Cheng, Qin, Lin, Hu, Yang, Ding, Sun, Xie,
  Liu, and Sun}]{su-etal-2023-exploring}
Yusheng Su, Chi-Min Chan, Jiali Cheng, Yujia Qin, Yankai Lin, Shengding Hu,
  Zonghan Yang, Ning Ding, Xingzhi Sun, Guotong Xie, Zhiyuan Liu, and Maosong
  Sun. 2023.
\newblock \href {https://doi.org/10.18653/v1/2023.emnlp-main.931} {Exploring
  the impact of model scaling on parameter-efficient tuning}.
\newblock In \emph{Proceedings of the 2023 Conference on Empirical Methods in
  Natural Language Processing}, pages 15062--15078, Singapore. Association for
  Computational Linguistics.

\bibitem[{Sun et~al.(2023)Sun, Zhang, Yan, Gao, Ong, Chen, and
  Su}]{sun-etal-2023-battle}
Shuo Sun, Yuchen Zhang, Jiahuan Yan, Yuze Gao, Donovan Ong, Bin Chen, and Jian
  Su. 2023.
\newblock \href {https://doi.org/10.18653/v1/2023.findings-emnlp.750} {Battle
  of the large language models: Dolly vs {LL}a{MA} vs vicuna vs guanaco vs bard
  vs {C}hat{GPT} - a text-to-{SQL} parsing comparison}.
\newblock In \emph{Findings of the Association for Computational Linguistics:
  EMNLP 2023}, pages 11225--11238, Singapore. Association for Computational
  Linguistics.

\bibitem[{Tabibian et~al.(2019)Tabibian, Upadhyay, De, Zarezade, Sch{\"o}lkopf,
  and Gomez-Rodriguez}]{Tabibian2019-ic}
Behzad Tabibian, Utkarsh Upadhyay, Abir De, Ali Zarezade, Bernhard
  Sch{\"o}lkopf, and Manuel Gomez-Rodriguez. 2019.
\newblock \href {https://www.pnas.org/doi/10.1073/pnas.1815156116} {Enhancing
  human learning via spaced repetition optimization}.
\newblock \emph{Proc. Natl. Acad. Sci. U. S. A.}, 116(10):3988--3993.

\bibitem[{Team et~al.(2023)Team, Anil, Borgeaud, Wu, Alayrac, Yu, Soricut,
  Schalkwyk, Dai, Hauth et~al.}]{team2023gemini}
Gemini Team, Rohan Anil, Sebastian Borgeaud, Yonghui Wu, Jean-Baptiste Alayrac,
  Jiahui Yu, Radu Soricut, Johan Schalkwyk, Andrew~M Dai, Anja Hauth, et~al.
  2023.
\newblock \href {https://arxiv.org/abs/2312.11805} {Gemini: a family of highly
  capable multimodal models}.
\newblock \emph{arXiv preprint arXiv:2312.11805}.

\bibitem[{Templin(1957)}]{templin1957certain}
Mildred~C. Templin. 1957.
\newblock \href {http://www.jstor.org/stable/10.5749/j.ctttv2st} {\emph{Certain
  Language Skills in Children: Their Development and Interrelationships}}, ned
  - new edition edition, volume~26.
\newblock University of Minnesota Press.

\bibitem[{Tjong Kim~Sang and
  Buchholz(2000)}]{tjong-kim-sang-buchholz-2000-introduction}
Erik~F. Tjong Kim~Sang and Sabine Buchholz. 2000.
\newblock \href {https://aclanthology.org/W00-0726/} {Introduction to the
  {C}o{NLL}-2000 shared task chunking}.
\newblock In \emph{Fourth Conference on Computational Natural Language Learning
  and the Second Learning Language in Logic Workshop}.

\bibitem[{Touvron et~al.(2023)Touvron, Lavril, Izacard, Martinet, Lachaux,
  Lacroix, Rozière, Goyal, Hambro, Azhar, Rodriguez, Joulin, Grave, and
  Lample}]{touvron2023llama}
Hugo Touvron, Thibaut Lavril, Gautier Izacard, Xavier Martinet, Marie-Anne
  Lachaux, Timothée Lacroix, Baptiste Rozière, Naman Goyal, Eric Hambro,
  Faisal Azhar, Aurelien Rodriguez, Armand Joulin, Edouard Grave, and Guillaume
  Lample. 2023.
\newblock \href {http://arxiv.org/abs/2302.13971} {Llama: Open and efficient
  foundation language models}.

\bibitem[{Vaswani et~al.(2017)Vaswani, Shazeer, Parmar, Uszkoreit, Jones,
  Gomez, Kaiser, and Polosukhin}]{NIPS2017_3f5ee243}
Ashish Vaswani, Noam Shazeer, Niki Parmar, Jakob Uszkoreit, Llion Jones,
  Aidan~N Gomez, \L~ukasz Kaiser, and Illia Polosukhin. 2017.
\newblock \href
  {https://proceedings.neurips.cc/paper_files/paper/2017/file/3f5ee243547dee91fbd053c1c4a845aa-Paper.pdf}
  {Attention is all you need}.
\newblock In \emph{Advances in Neural Information Processing Systems}.

\bibitem[{Voita et~al.(2019)Voita, Talbot, Moiseev, Sennrich, and
  Titov}]{voita-etal-2019-analyzing}
Elena Voita, David Talbot, Fedor Moiseev, Rico Sennrich, and Ivan Titov. 2019.
\newblock \href {https://doi.org/10.18653/v1/P19-1580} {Analyzing multi-head
  self-attention: Specialized heads do the heavy lifting, the rest can be
  pruned}.
\newblock In \emph{Proceedings of the 57th Annual Meeting of the Association
  for Computational Linguistics}, pages 5797--5808, Florence, Italy.
  Association for Computational Linguistics.

\bibitem[{Warstadt et~al.(2020)Warstadt, Parrish, Liu, Mohananey, Peng, Wang,
  and Bowman}]{warstadt-etal-2020-blimp-benchmark}
Alex Warstadt, Alicia Parrish, Haokun Liu, Anhad Mohananey, Wei Peng, Sheng-Fu
  Wang, and Samuel~R. Bowman. 2020.
\newblock \href {https://doi.org/10.1162/tacl_a_00321} {{BL}i{MP}: The
  benchmark of linguistic minimal pairs for {E}nglish}.
\newblock \emph{Transactions of the Association for Computational Linguistics},
  8:377--392.

\bibitem[{Wei et~al.(2021)Wei, Finn, Templeton, Wheatley, and
  Vosoughi}]{wei-etal-2021-linguistic}
Jason Wei, Kelly Finn, Emma Templeton, Thalia Wheatley, and Soroush Vosoughi.
  2021.
\newblock \href {https://doi.org/10.18653/v1/2021.naacl-main.352} {Linguistic
  complexity loss in text-based therapy}.
\newblock In \emph{Proceedings of the 2021 Conference of the North American
  Chapter of the Association for Computational Linguistics: Human Language
  Technologies}, pages 4450--4459, Online. Association for Computational
  Linguistics.

\bibitem[{Wei et~al.(2022)Wei, Wang, Schuurmans, Bosma, brian ichter, Xia, Chi,
  Le, and Zhou}]{wei2022chain}
Jason Wei, Xuezhi Wang, Dale Schuurmans, Maarten Bosma, brian ichter, Fei Xia,
  Ed~H. Chi, Quoc~V Le, and Denny Zhou. 2022.
\newblock \href {https://openreview.net/forum?id=_VjQlMeSB_J} {Chain of thought
  prompting elicits reasoning in large language models}.
\newblock In \emph{Advances in Neural Information Processing Systems}.

\bibitem[{Weller et~al.(2020)Weller, Lourie, Gardner, and
  Peters}]{weller-etal-2020-learning}
Orion Weller, Nicholas Lourie, Matt Gardner, and Matthew~E. Peters. 2020.
\newblock \href {https://doi.org/10.18653/v1/2020.emnlp-main.105} {Learning
  from task descriptions}.
\newblock In \emph{Proceedings of the 2020 Conference on Empirical Methods in
  Natural Language Processing (EMNLP)}, pages 1361--1375, Online. Association
  for Computational Linguistics.

\bibitem[{Wolfe-Quintero et~al.(1998)Wolfe-Quintero, Inagaki, and
  Kim}]{wolfquintero1998}
Kate Wolfe-Quintero, Shunji Inagaki, and Hae-Young Kim. 1998.
\newblock \href {https://nflrc.hawaii.edu/publications/view/tr17/} {Second
  language development in writing: Measures of fluency, accuracy, and
  complexity}.
\newblock \emph{Honolulu, HI: University of Hawai'i, Second Language Teaching
  \& Curriculum Center}.

\bibitem[{Xia et~al.(2016)Xia, Kochmar, and Briscoe}]{xia-etal-2016-text}
Menglin Xia, Ekaterina Kochmar, and Ted Briscoe. 2016.
\newblock \href {https://doi.org/10.18653/v1/W16-0502} {Text readability
  assessment for second language learners}.
\newblock In \emph{Proceedings of the 11th Workshop on Innovative Use of {NLP}
  for Building Educational Applications}, pages 12--22, San Diego, CA.
  Association for Computational Linguistics.

\bibitem[{Xu et~al.(2015)Xu, Callison-Burch, and
  Napoles}]{xu-etal-2015-problems}
Wei Xu, Chris Callison-Burch, and Courtney Napoles. 2015.
\newblock \href {https://doi.org/10.1162/tacl_a_00139} {Problems in current
  text simplification research: New data can help}.
\newblock \emph{Transactions of the Association for Computational Linguistics},
  3:283--297.

\bibitem[{Yamaguchi et~al.(2021)Yamaguchi, Chrysostomou, Margatina, and
  Aletras}]{yamaguchi-etal-2021-frustratingly}
Atsuki Yamaguchi, George Chrysostomou, Katerina Margatina, and Nikolaos
  Aletras. 2021.
\newblock \href {https://doi.org/10.18653/v1/2021.emnlp-main.249}
  {Frustratingly simple pretraining alternatives to masked language modeling}.
\newblock In \emph{Proceedings of the 2021 Conference on Empirical Methods in
  Natural Language Processing}, pages 3116--3125, Online and Punta Cana,
  Dominican Republic. Association for Computational Linguistics.

\bibitem[{Yang and Tu(2022)}]{yang-tu-2022-bottom}
Songlin Yang and Kewei Tu. 2022.
\newblock \href {https://doi.org/10.18653/v1/2022.acl-long.171} {Bottom-up
  constituency parsing and nested named entity recognition with pointer
  networks}.
\newblock In \emph{Proceedings of the 60th Annual Meeting of the Association
  for Computational Linguistics (Volume 1: Long Papers)}, pages 2403--2416,
  Dublin, Ireland. Association for Computational Linguistics.

\bibitem[{Yannakoudakis et~al.(2011)Yannakoudakis, Briscoe, and
  Medlock}]{yannakoudakis-etal-2011-new}
Helen Yannakoudakis, Ted Briscoe, and Ben Medlock. 2011.
\newblock \href {https://aclanthology.org/P11-1019/} {A new dataset and method
  for automatically grading {ESOL} texts}.
\newblock In \emph{Proceedings of the 49th Annual Meeting of the Association
  for Computational Linguistics: Human Language Technologies}, pages 180--189,
  Portland, Oregon, USA. Association for Computational Linguistics.

\bibitem[{Yao et~al.(2023)Yao, Zhao, Yu, Du, Shafran, Narasimhan, and
  Cao}]{yao2023react}
Shunyu Yao, Jeffrey Zhao, Dian Yu, Nan Du, Izhak Shafran, Karthik~R Narasimhan,
  and Yuan Cao. 2023.
\newblock \href {https://openreview.net/forum?id=WE_vluYUL-X} {React:
  Synergizing reasoning and acting in language models}.
\newblock In \emph{The Eleventh International Conference on Learning
  Representations}.

\bibitem[{Yu et~al.(2023)Yu, Buchanan, Pai, Chu, Wu, Tong, Haeffele, and
  Ma}]{yu2023white}
Yaodong Yu, Sam Buchanan, Druv Pai, Tianzhe Chu, Ziyang Wu, Shengbang Tong,
  Benjamin~David Haeffele, and Yi~Ma. 2023.
\newblock \href {https://openreview.net/forum?id=THfl8hdVxH} {White-box
  transformers via sparse rate reduction}.
\newblock In \emph{Thirty-seventh Conference on Neural Information Processing
  Systems}.

\bibitem[{Zareva et~al.(2005)Zareva, Schwanenflugel, and
  Nikolova}]{zareva2005relationship}
Alla Zareva, Paula Schwanenflugel, and Yordanka Nikolova. 2005.
\newblock \href
  {https://www.cambridge.org/core/journals/studies-in-second-language-acquisition/article/relationship-between-lexical-competence-and-language-proficiency-variable-sensitivity/5B72A4A12AF25DE8F803846C0ED13D78}
  {Relationship between lexical competence and language proficiency: Variable
  sensitivity}.
\newblock \emph{Studies in Second Language Acquisition}, 27(4):567--595.

\bibitem[{Zheng et~al.(2023)Zheng, Chiang, Sheng, Zhuang, Wu, Zhuang, Lin, Li,
  Li, Xing, Zhang, Gonzalez, and Stoica}]{zheng2023judging}
Lianmin Zheng, Wei-Lin Chiang, Ying Sheng, Siyuan Zhuang, Zhanghao Wu, Yonghao
  Zhuang, Zi~Lin, Zhuohan Li, Dacheng Li, Eric.~P Xing, Hao Zhang, Joseph~E.
  Gonzalez, and Ion Stoica. 2023.
\newblock \href {http://arxiv.org/abs/2306.05685} {Judging llm-as-a-judge with
  mt-bench and chatbot arena}.

\bibitem[{Zhou et~al.(2023)Zhou, Liu, Xu, Iyer, Sun, Mao, Ma, Efrat, Yu, Yu
  et~al.}]{zhou2023lima}
Chunting Zhou, Pengfei Liu, Puxin Xu, Srini Iyer, Jiao Sun, Yuning Mao, Xuezhe
  Ma, Avia Efrat, Ping Yu, Lili Yu, et~al. 2023.
\newblock Lima: Less is more for alignment.
\newblock \emph{arXiv preprint arXiv:2305.11206}.

\bibitem[{Ziems et~al.(2024)Ziems, Held, Dwivedi-Yu, and
  Yang}]{ziems-etal-2024-measuring}
Caleb Ziems, William Held, Jane Dwivedi-Yu, and Diyi Yang. 2024.
\newblock \href {https://doi.org/10.18653/v1/2024.findings-acl.763} {Measuring
  and addressing indexical bias in information retrieval}.
\newblock In \emph{Findings of the Association for Computational Linguistics:
  ACL 2024}, pages 12860--12877, Bangkok, Thailand. Association for
  Computational Linguistics.

\end{thebibliography}
\bibliographystyle{acl_natbib}

\newpage
\appendix
\newpage
\newpage
\section{GPT's Knowledge on Target tasks}\label{sec:gpt_knowledge}

\begin{figure}[t]
    \centering
    \includegraphics[scale=.5]{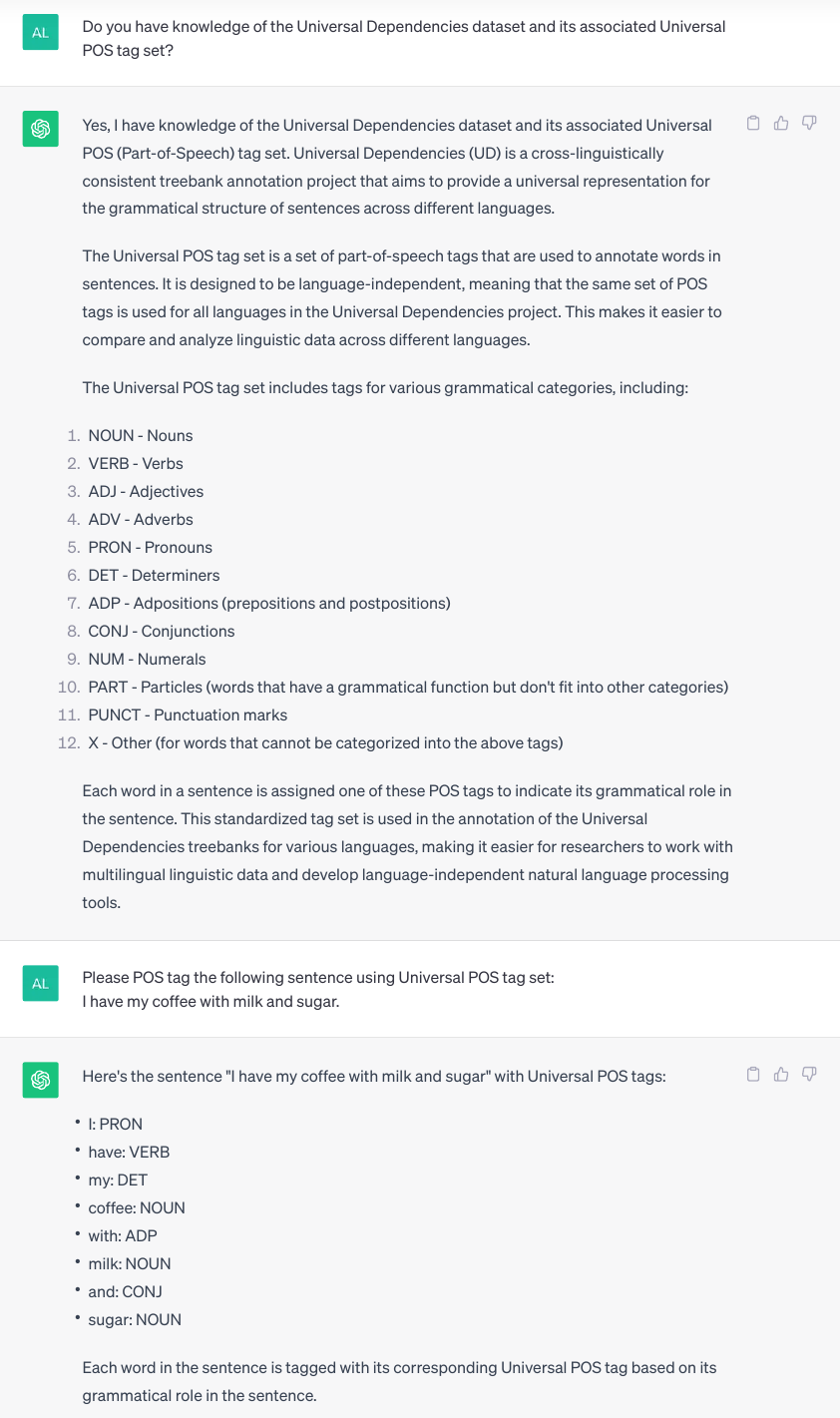}
    \caption{GPT's responses to our questions about universal dependencies dataset and its associated universal POS tag set.}
\label{fig:gpt_pos}
\end{figure}

\begin{figure*}
  \begin{minipage}{0.4\textwidth} 
    \centering
    \includegraphics[width=\linewidth]{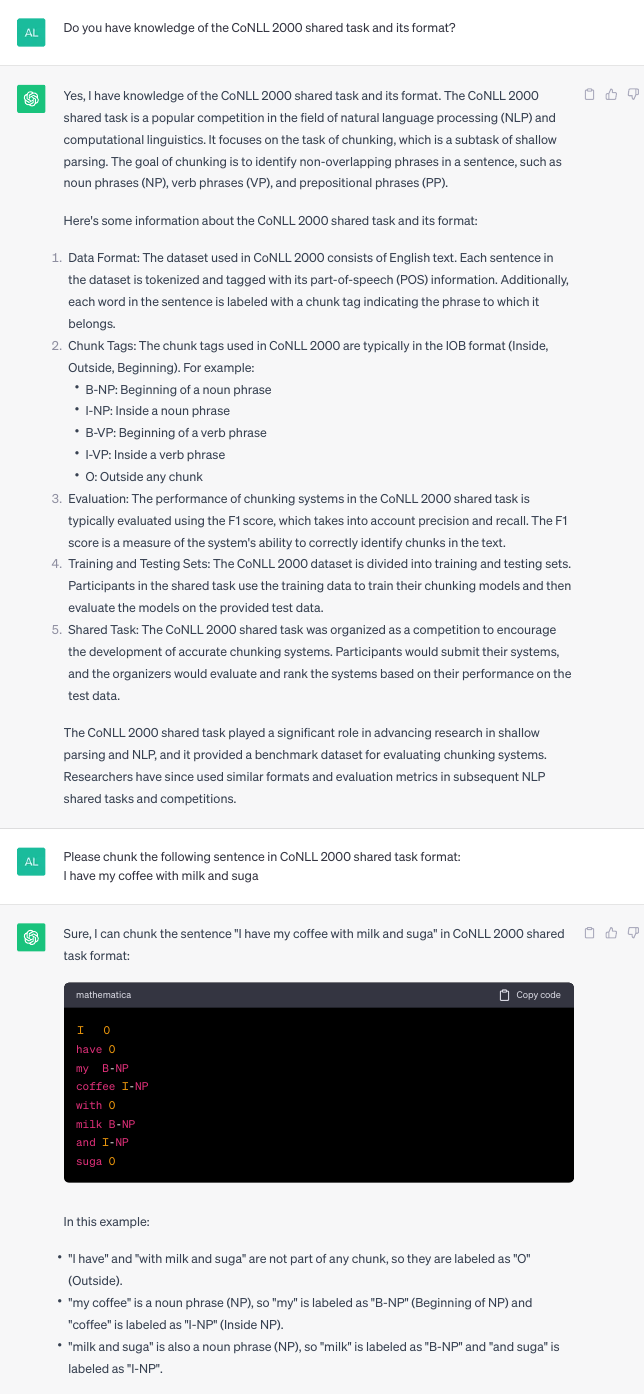}
    \caption{GPT's responses to our questions about the CoNLL 2000 shared task and its format.}
    \label{fig:gpt_coNLL}
  \end{minipage}  
  \hfill
  \begin{minipage}{0.4\textwidth} 
    \centering
    \includegraphics[width=\linewidth]{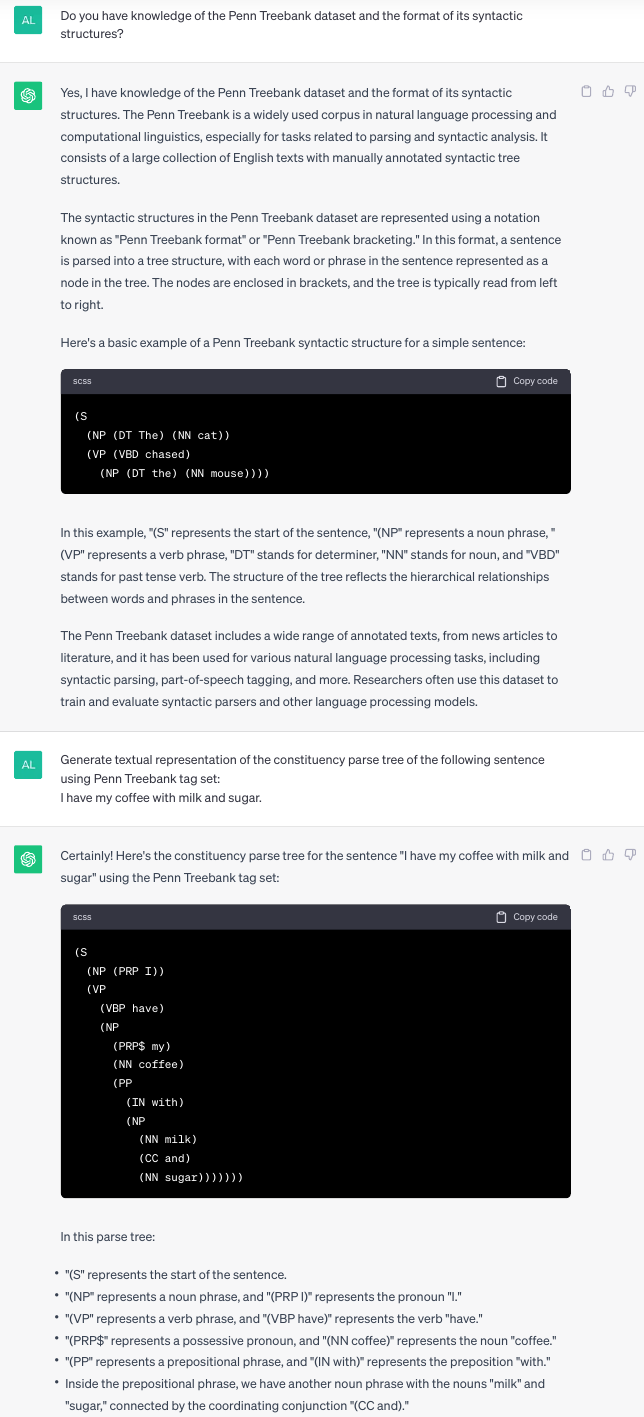}
    \caption{GPT's responses to our questions about the Penn Treebank dataset and the format of its syntactic structures.}
    \label{fig:gpt_parsing}
  \end{minipage}
\end{figure*}

As illustrated in Figures~\ref{fig:gpt_pos}--\ref{fig:gpt_parsing}, we asked relevant questions from GPT 3.5 about the target linguistic tasks of this study. The responses clearly indicate that GPT 3.5 have knowledge about the universal dependencies dataset and the universal POS tag set, the CoNLL 2000 shared task and its format, and the Penn Treebank dataset and the format of its syntactic structures.

% \begin{figure}[t]
%     \centering
%     \includegraphics[scale=.6]{figure/gpt_coNLL.pdf}
%     \caption{GPT's responses to our questions about the CoNLL 2000 shared task and its format.}
% \label{fig:gpt_coNLL}
% \end{figure}

% \begin{figure}[t]
%     \centering
%     \includegraphics[scale=.6]{figure/gpt_parsing.pdf}
%     \caption{GPT's responses to our questions about the Penn Treebank dataset and the format of its syntactic structures.}
% \label{fig:gpt_parsing}
% \end{figure}

\section{Detailed Results}\label{sec:more_result}

\subsection{Performance of all LLMs}
We present the performances of all LLMs in Figure~\ref{fig:qp_gemini}--\ref{fig:qp_mistral-7b}.

\subsection{Performance across linguistic complexity spectrum}
\begin{figure*}[h]
    \centering
    \includegraphics[width=0.98\textwidth]{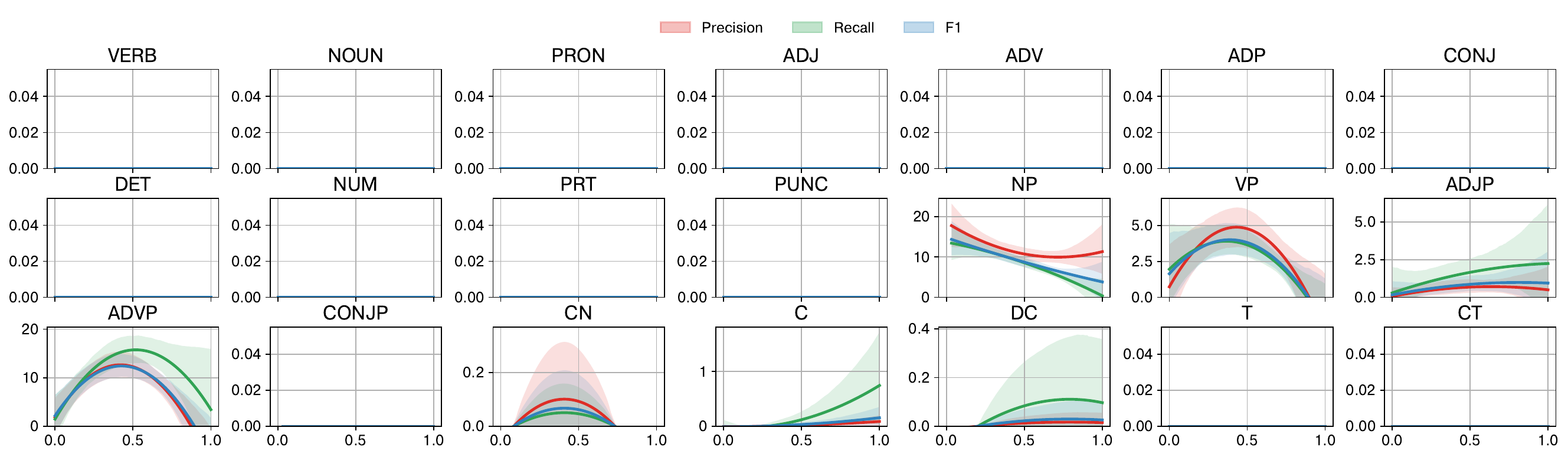}
    \caption{Performance of Gemini with respect to linguistic complexity.}
\label{fig:qp_gemini}
\end{figure*}

\begin{figure*}[h]
    \centering
    \includegraphics[width=0.9\textwidth]{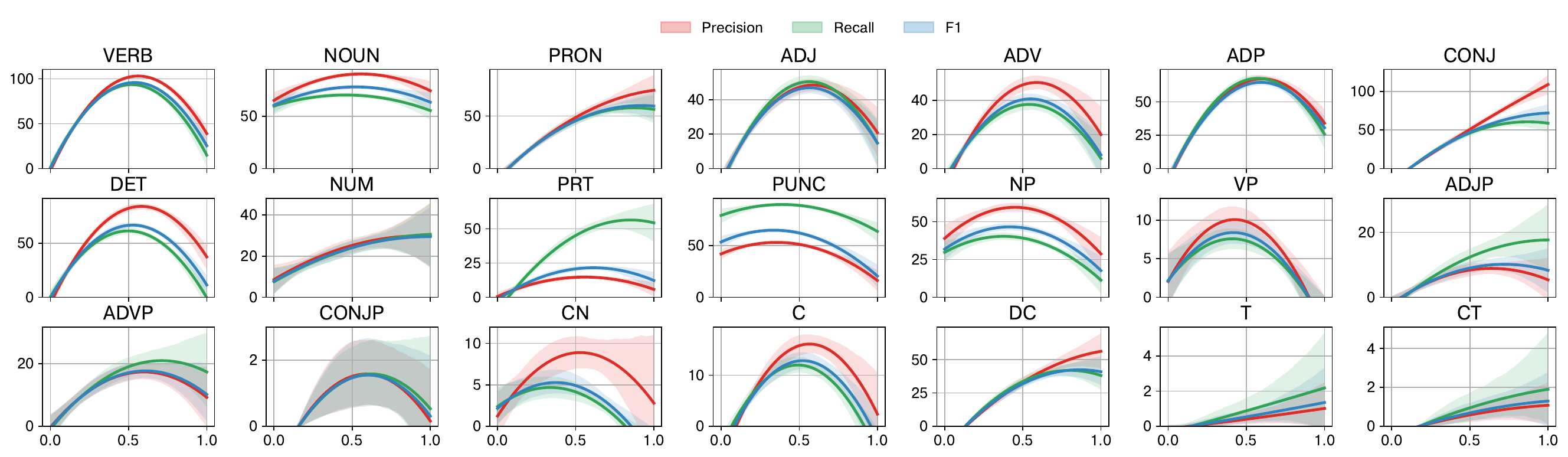}
    \caption{Performance of LLaMA3-70b with respect to linguistic complexity.}
\label{fig:qp_llama3-70b}
\end{figure*}

\begin{figure*}[h]
    \centering
    \includegraphics[width=0.98\textwidth]{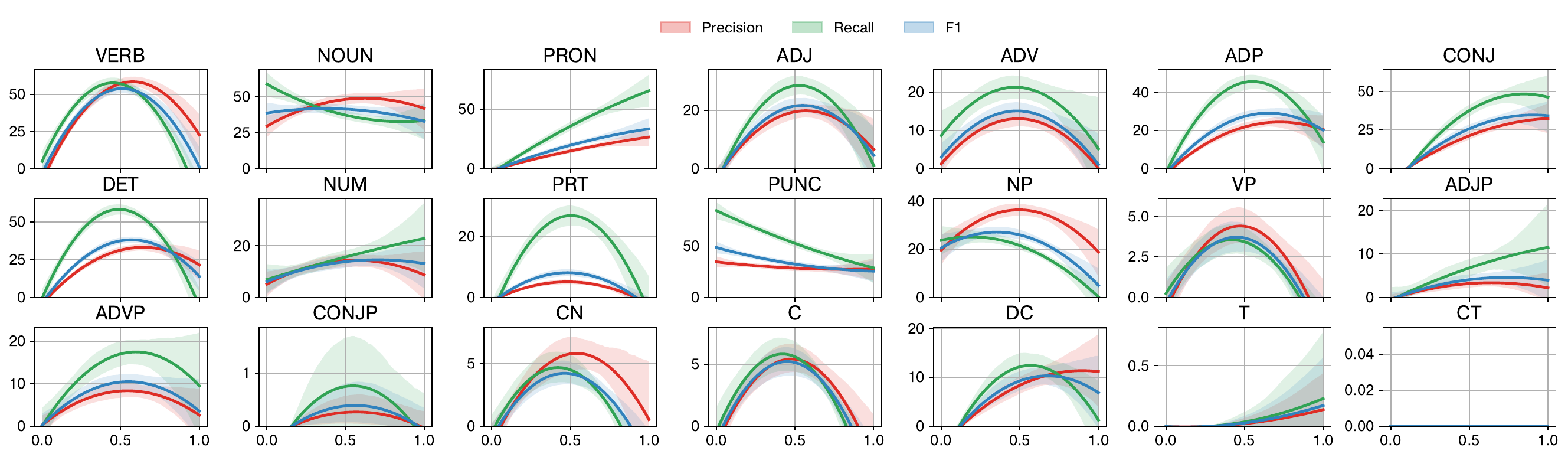}
    \caption{Performance of LLaMA2-70B with respect to linguistic complexity.}
\label{fig:qp_llama2-70b}
\end{figure*}

\begin{figure*}[h]
    \centering
    \includegraphics[width=0.98\textwidth]{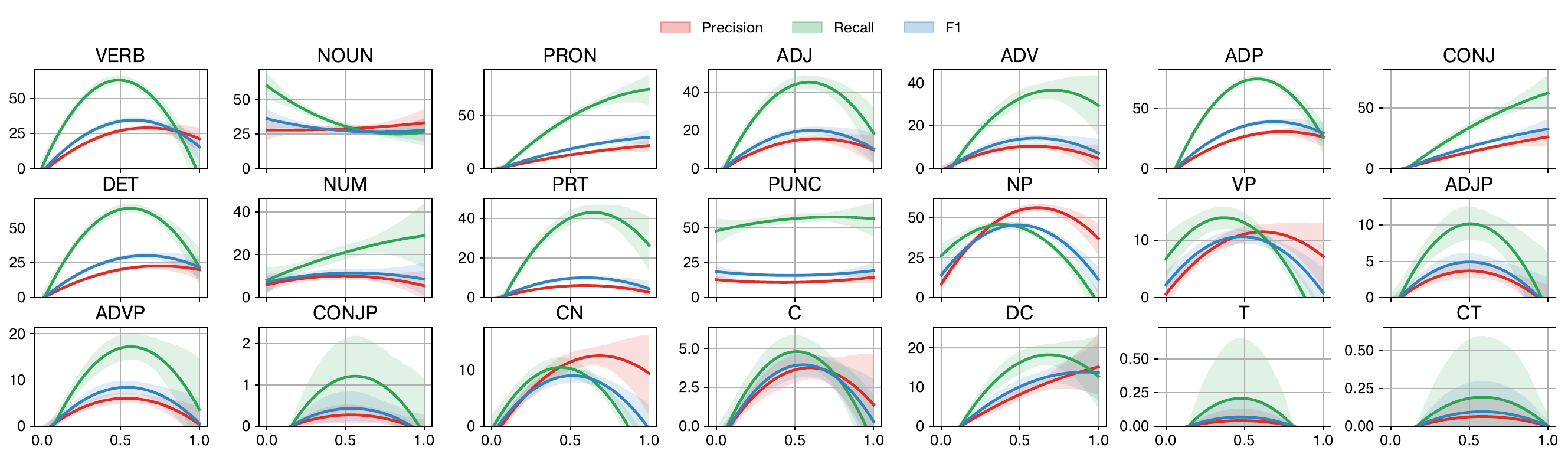}
    \caption{Performance of Mixtral-8x7B with respect to linguistic complexity.}
\label{fig:qp_mistral-8x7b}
\end{figure*}

\begin{figure*}[h]
    \centering
    \includegraphics[width=0.98\textwidth]{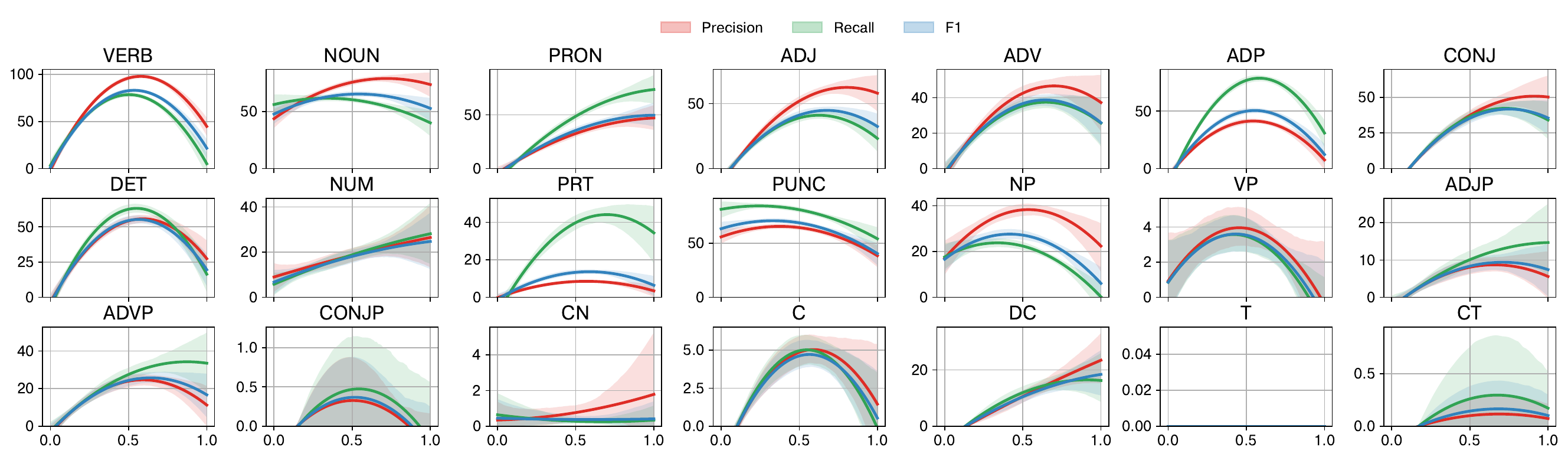}
    \caption{Performance of LLaMA3-8B with respect to linguistic complexity.}
\label{fig:qp_llama3-8b}
\end{figure*}

\begin{figure*}[h]
    \centering
    \includegraphics[width=0.98\textwidth]{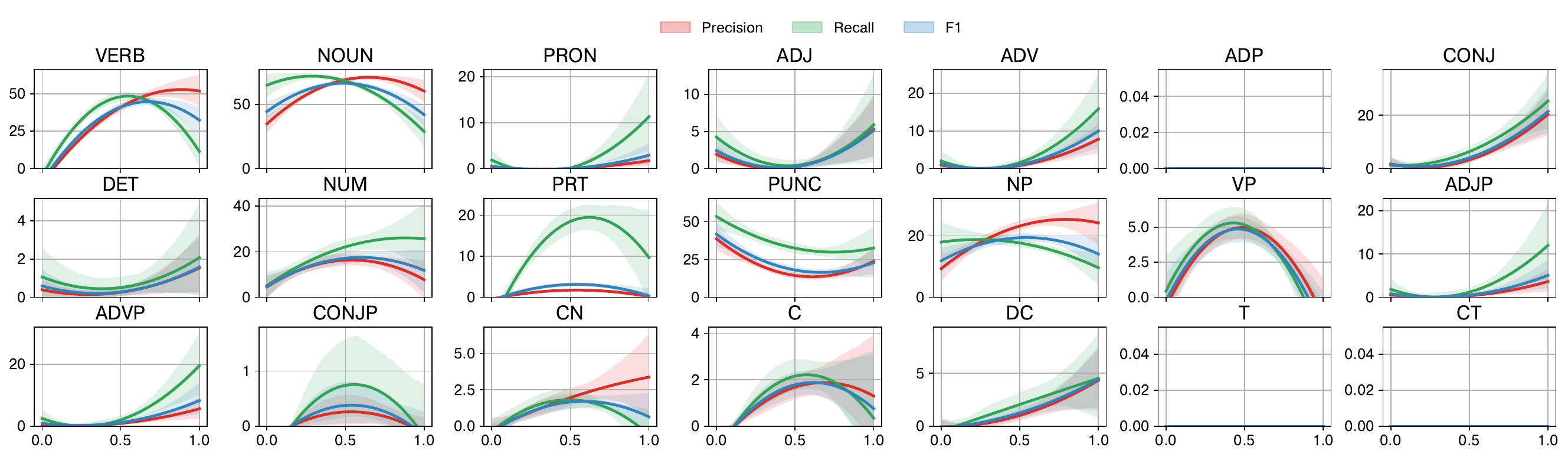}
    \caption{Performance of LLaMA2-7B with respect to linguistic complexity.}
\label{fig:qp_llama2-7b}
\end{figure*}

\begin{figure*}[h]
    \centering
    \includegraphics[width=0.98\textwidth]{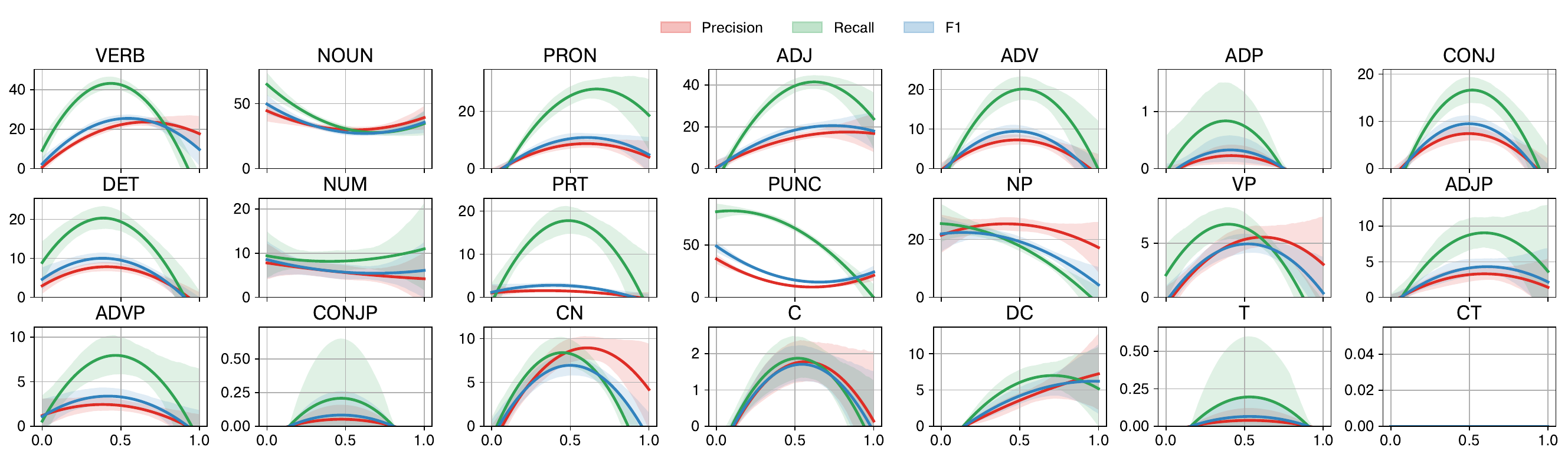}
    \caption{Performance of Mistral-7B with respect to linguistic complexity.}
\label{fig:qp_mistral-7b}
\end{figure*}

We present the performance-sample complexity plot for all models in Figure~\ref{fig:qp_gemini}-Figure\ref{fig:qp_mistral-7b}. These results suggest that LLM's linguistic capability largely depends on the linguistic complexity of their inputs, and existing work may have over-estimated the linguistic capability of LLMs. However, it's crucial to note that linguistic complexity as understood by humans and machines may differ significantly. Indeed, there are particular linguistic structures where model performance surprisingly escalates with increasing complexity. Understanding the reason for such model behaviors require understanding their learning dynamics and is the subject of our future work.

\section{Linguistic indices}\label{sec:app}

Table~\ref{tab:ling_ind} presents the 45 linguistic indices in our study. 
% We also compute the correlation coefficient between all pairs of indices, illustrated in Fig.~\ref{fig:corr}.

\begin{table}[h]
\small
\centering
\begin{tabular}{l|l|l}
\toprule
    Granularity & Name & Notation \\
    \midrule
    Word & Nouns & Num\_NN \\
    Word & Verbs & Num\_VB \\
    Word & Adjectives & Num\_JJ \\
    Word & Adverbs & Num\_RB \\
    Word & Prepositions/Subordinates & Num\_IN \\
    Word & Coordinating Conjunction & Num\_CC \\
    Word & Determiner & Num\_DT \\
    \midrule
    Phrase & Noun Phrases & Num\_NP \\
    Phrase & Verb Phrases & Num\_VP \\
    Phrase & Adjective Phrases & Num\_ADJP \\
    Phrase & Adverb Phrases & Num\_ADVP \\
    Phrase & Preposition Phrases & Num\_PP \\
    Phrase & Conjunction Phrases & Num\_CONJP \\
    Phrase & Quantitative Phrases & Num\_QP \\
    Phrase & Complex Nominal & Num\_CN \\
    \midrule
    Sentence & T-Units & Num\_T \\
    Sentence & Complex T-Units & Num\_CT \\
    Sentence & Clause & Num\_C \\
    Sentence & Dependent Clause & Num\_DC \\
    Sentence & Fragment Clause & Num\_FC \\
    % \cmidrule{2-5}
    % \cmidrule{2-5}
    % Complexity & Sentence & Mean length of clause & MLC \\
    % Complexity & Sentence & Mean length of sentence & MLS \\
    % Complexity & Sentence & Mean length of T-Unit & MLT \\
    % Complexity & Sentence & Sentence complexity ratio & C/S \\
    % Complexity & Sentence & T-unit complexity ratio & C/T \\
    % Complexity & Sentence & Complex T-unit proportion & CT/T \\
    % Complexity & Sentence & Dependent Clause proportion & DC/C \\
    % Complexity & Sentence & Dependent Clause to T-Unit ratio & DC/T \\
    % Complexity & Sentence & Sentence coordination ratio & T/S \\
    % \cmidrule{2-5}
    % Complexity & Phrase \& Sentence & Coordinate phrases to clause ratio & CP/C \\
    % Complexity & Phrase \& Sentence & Coordinate phrases to T-Unit ratio & CP/T \\
    % Complexity & Phrase \& Sentence & Complex nominals to clause ratio & CN/C \\
    % Complexity & Phrase \& Sentence & Complex nominals to T-unit ratio & CN/T \\
    % Complexity & Phrase \& Sentence & Verb phrases to T-unit ratio & VP/T \\
    % \cmidrule{2-5}
    % Complexity & Word \& Sentence & Flesch Kincaid Grade Level & FleschG\_S \\
    % Complexity & Word \& Sentence & Automated Readability Index & AutoRea\_S \\
    % Complexity & Word \& Sentence & Coleman Liau Readability Score & ColeLia\_S \\
    % Complexity & Word \& Sentence & Smog Index & SmogInd\_S \\
    % Complexity & Word \& Sentence & Gunning Fog Count Score & Gunning\_S \\
    % Complexity & Word \& Sentence & Linsear Write Formula Score & LinseaW\_S \\
\bottomrule
\end{tabular}
\caption{Linguistic indices we use in the study.}
\label{tab:ling_ind}
\end{table}

% Type–Token Ratio TTR T/N
% Mean Segmental TTR (50) MSTTR–50 Mean TTR of all 50-word segments
% Corrected TTR CTTR T/√2N
% Root TTR RTTR T/√N
% Bilogarithmic TTR LogTTR LogT/Log N
% Uber Index Uber Log 2N/Log(N/T)
% D Measure D Based on D in Equation (1)
% Lexical Word Variation LV Tlex/Nlex
% Verb Variation-I VV1 Tverb /Nverb
% Squared VV1 SVV1 Tv2
% erb /Nverb
% Corrected VV1 CVV1 Tverb /√2Nverb
% Verb Variation-II VV2 Tverb /Nlex
% Noun Variation NV Tnoun/Nlex
% Adjective Variation AdjV Tadj /Nlex
% Adverb Variation AdvV Tadv /Nlex
% Modifier Variation ModV (Tadj + Tadv )/Nlex
% TTR & Type–token ratio &  Lexical & Sophistication & \\

\end{document}